\documentclass[10pt]{article}

\usepackage{times}
\usepackage{amsmath}
\usepackage{amssymb}
\usepackage{cite}
\usepackage{bm}
\usepackage{float}
\usepackage{url}
\usepackage[ruled]{algorithm}
\usepackage{algorithmicx}
\usepackage{algpseudocode}
\usepackage{graphicx}
\usepackage{subfigure}
\usepackage[small]{caption}
\usepackage{color}
\usepackage{graphics}
\usepackage{graphicx,epsfig,psfrag,pstricks}
\usepackage{multirow}
\usepackage{array}

\newcommand{\boldA}{\mathbf{A}}

\newcommand{\boldz}{\mathbf{z}}

\newcommand{\boldI}{\mathbf{I}}

\title{A Tucker decomposition process for probabilistic modeling of diffusion magnetic resonance imaging}
\author{Hern\'an Dar\'io Vargas Cardona, Mauricio A. \'Alvarez and Alvaro A. Orozco\\
{\small \emph{Faculty of Engineering, Universidad Tecnol{\'o}gica de Pereira, Colombia, 660003.}}\\}
\date{}

\begin{document}
\maketitle

\begin{abstract}
Diffusion magnetic resonance imaging (dMRI) is an emerging medical technique used for describing water diffusion in an organic tissue. Typically, rank-2 tensors quantify this diffusion. From this quantification, it is possible to calculate relevant scalar measures (i.e. fractional anisotropy and mean diffusivity) employed in clinical diagnosis of neurological diseases. Nonetheless, 2nd-order tensors fail to represent complex tissue structures like crossing fibers. To overcome this limitation, several researchers proposed a diffusion representation with higher order tensors (HOT), specifically 4th and 6th orders. However, the current acquisition protocols of dMRI data allow images with a spatial resolution between 1 $mm^3$ and 2 $mm^3$. This voxel size is much smaller than tissue structures. Therefore, several clinical procedures derived from dMRI may be inaccurate. Interpolation has been used to enhance resolution of dMRI in a tensorial space. Most interpolation methods are valid only for rank-2 tensors and a generalization for HOT data is missing. In this work, we propose a novel stochastic process called Tucker decomposition process (TDP) for performing HOT data interpolation. Our model is based on the Tucker decomposition and Gaussian processes as parameters of the TDP. We test the TDP in 2nd, 4th and 6th rank HOT fields. For rank-2 tensors, we compare against direct interpolation, log-Euclidean approach and Generalized Wishart processes. For rank-4 and rank-6 tensors we compare against direct interpolation. Results obtained show that TDP interpolates accurately the HOT fields and generalizes to any rank.
\end{abstract}

\section{Introduction}

Diffusion magnetic resonance imaging (dMRI) is a medical technique that non-invasively  describes water diffusion in organic tissue. The first attempt to represent this physical phenomena was the Gaussian model proposed by \cite{Basser1993,Basser1994}, where symmetric and positive definite tensors of rank-2 quantify the direction and orientation of diffusion. From this quantification, it is possible to compute relevant physiological information (i.e. Fractional anisotropy and mean diffusivity) employed in neuro-degenerative diseases: deep brain stimulation over Parkinson's disease patients \cite{Butson2007}, trauma \cite{Ptak2003}, multiple sclerosis \cite{Hasan2005}, meningitis \cite{Nath2007}, among others. Nevertheless, rank-2 tensors fail to represent accurately some complex tissue structures such as: white matter fiber bundles, crossing fibers and bifurcated fibers \cite{Ozarslan2003,Mori1999}.

To address these limitations in dMRI, several researchers \cite{Ozarslan2003,Liu2004,Moakher2008,Barmpoutis2010} proposed higher order tensor (HOT) models in order to describe diffusion inside complex tissue structures. These models demonstrated power and flexibility to analyze dMRI data. However, the current acquisition protocols of dMRI data allow images with a spatial resolution in a range from 1 $mm^3$ and 2 $mm^3$. This voxel size is much smaller than tissue fibers. Therefore, clinical procedures derived from dMRI may be inaccurate. Interpolation of HOT fields is a feasible methodology to enhance spatial resolution in dMRI and takes relevance in the reconstruction of tissue fiber bundles (this procedure is known as tractography). Also, in image registration algorithms or any application where it is required to estimate data among nearby tensors of the field \cite{Yassine2009}.

A considerable number of methods for dMRI data interpolation have been proposed in the literature. Direct linear interpolation \cite{Pajevic2002}, log-Euclidean space \cite{Arsigny2006}, b-splines \cite{Barmpoutis2007}, Riemannian manifolds \cite{Pennec2006,Flecther2007}, feature-based framework \cite{Yang2012}, geodesic loxodromes \cite{Kindlmann2007} and generalized Wishart processes \cite{Vargas2015}. They have different shortcomings. For example, \cite{Pajevic2002} does not ensure positive definite tensors, and \cite{Arsigny2006,Pennec2006,Flecther2007} are highly affected by the intrinsic Rician noise added in dMRI during acquisition. However, the most important limitation for all the  approaches mentioned, is that they are only valid for rank-2 tensors, and only the direct interpolation can be generalized to HOT fields. Regarding HOT field interpolation, the authors of \cite{Yassine2008,Yassine2009} developed a method based on tensor subdivision and minimization of two properties (curl and divergence) of the field for interpolation of 4th-order tensors. However, \cite{Yassine2008,Yassine2009} only reported outcomes for rank-4 tensors fields, and the method is not clear an extension to any order. Also, the authors of \cite{Astola2009,Astola2011} proposed an approach to perform probabilistic tractography in HOT data. This methodology is based on Finsler geometry. They developed the geometric generalization appropriate for multi-fiber analysis and demonstrated that a HOT field belongs to a Finsler manifold. Although, the Finsler geometry model is able to perform tractography in tensors fields of any order, it has not been established as a super-resolution method.

To the best of our knowledge, there is not a generalized methodology for super-resolution in HOT fields (no matter the rank), that retains all mandatory constraints for tensorial representation of dMRI. In this work, we propose a novel methodology to perform interpolation in HOT fields of any order. We develop a new stochastic process called Tucker decomposition process (TDP). Our model is based on the Tucker decomposition of tensors and a set of Gaussian processes that modulate the parameters of the decomposition as function of a two-dimensional space variable. We test the TDP in 2nd, 4th and 6th rank HOT fields.  For rank-2 tensors, we compare against direct interpolation \cite{Pajevic2002}, log-Euclidean approach \cite{Arsigny2006} and Generalized Wishart processes \cite{Vargas2015}. For rank-4 and rank-6 tensors we compare against direct interpolation. Results obtained show that TDP interpolates accurately the HOT fields, and generalizes to any rank. Also, the TDP keeps the mandatory constraints such as positive definite tensors. 

\section{Materials and methods}
In this section, we first define higher order tensors in dMRI. We then introduce the Tucker decomposition of a tensor. Next, we describe the prior that we use to represent a field of higher order tensors by combining the Tucker decomposition with Gaussian processes. Bayesian inference for the probabilistic model is then discussed. Finally, the experimental setup is detailed. 

\subsection{Higher Order Tensors and dMRI}

A tensor is a geometric or physical object specified by a set of coefficients $\mathcal{T}_{ijk...m}$ of a multi-linear form $\phi=\phi(\mathbf{x},\mathbf{y},\mathbf{z},...,\mathbf{w})$ of $l$ vector arguments $\mathbf{x},\mathbf{y},\mathbf{z},...,\mathbf{w}$ written in some orthonormal basis. 
The number $l$ is known as the order or rank of the tensor and each vector argument has an independent (may be different) dimensionality. A specific case is the Higher Order Tensor (HOT) employed for modeling diffusion of water particles in organic tissue. The HOT has a dimensionality $n=3$ for all arguments. A structured diffusion process through diffusion magnetic resonance imaging (dMRI) is defined by the generalized Stejskal-Tanner formula \cite{Ozarslan2003}:
\begin{equation}
	\log S_k=\log S_0 - b \times \sum_{i_1=1}^{3}\sum_{i_2=1}^{3}\cdots \sum_{i_l=1}^{3} \mathcal{D}_{i_1i_2...i_l}^{(l)}g_{i_1i_2...i_l}^{(l)},
	\label{gst}
\end{equation}
where $S_k$ is the $k_{th}$ dMRI acquired, in a particular input position, $S_0$ is the baseline image, $b$ is the diffusion coefficient, $l$ is the order of tensor, $g_{i_1i_2...i_l}$ is the direction of a gradient vector, and $\mathcal{D}_{i_1i_2...i_l}$ is the diffusion tensor. From equation \eqref{gst} it is possible to compute all elements of a tensor using a multi-linear regression \cite{Barmpoutis2010}, for all input locations in a dMRI. In dMRI, the diffusion function $\mathcal{D}(g)$ is defined as 
\begin{equation}
	\mathcal{D}(g)=\sum_{i_1=1}^{3}\sum_{i_2=1}^{3}\cdots \sum_{i_l=1}^{3} \mathcal{D}_{i_1i_2...i_l}g_{i_1i_2...i_l}, \nonumber
\end{equation}
The order $l$ must be strictly even, because if $l$ is odd implies that $\mathcal{D}(-g)=-\mathcal{D}(g)$, leading to non-positive definite tensors, that do not have physical interpretation in this context. For this reason, the rank of a higher order tensor (HOT) always must be even. A rank-$l$ tensor has $3^l$ elements. This number is very large for higher orders. But, total symmetry of HOT reduces significantly the number of unique components to $N_l=\frac{(l+1)(l+2)}{2}$.

\begin{figure}
	\centering
	\subfigure[]{\includegraphics[scale=0.3]{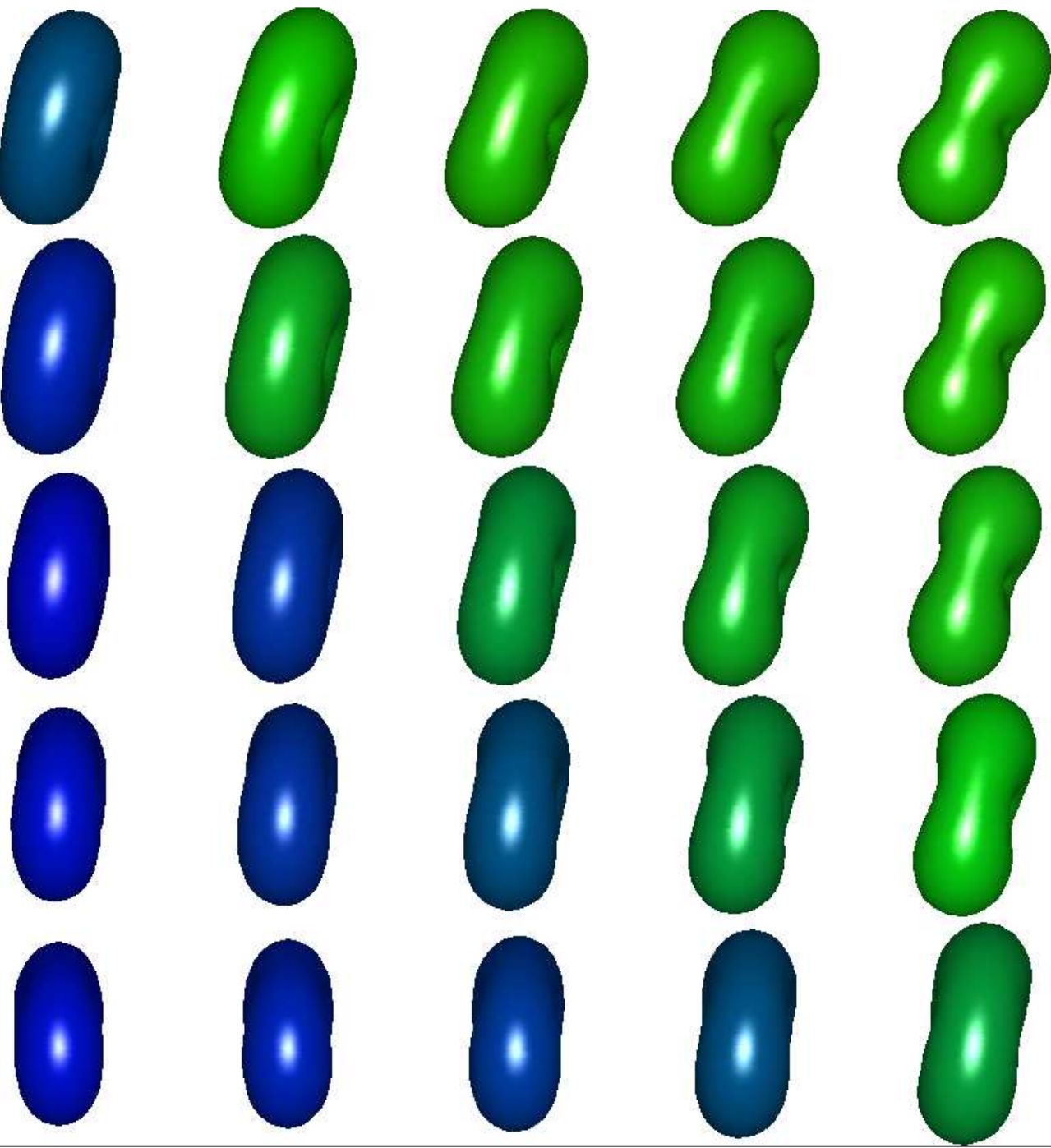}}\hspace{0.4cm}
	\subfigure[]{\includegraphics[scale=0.3]{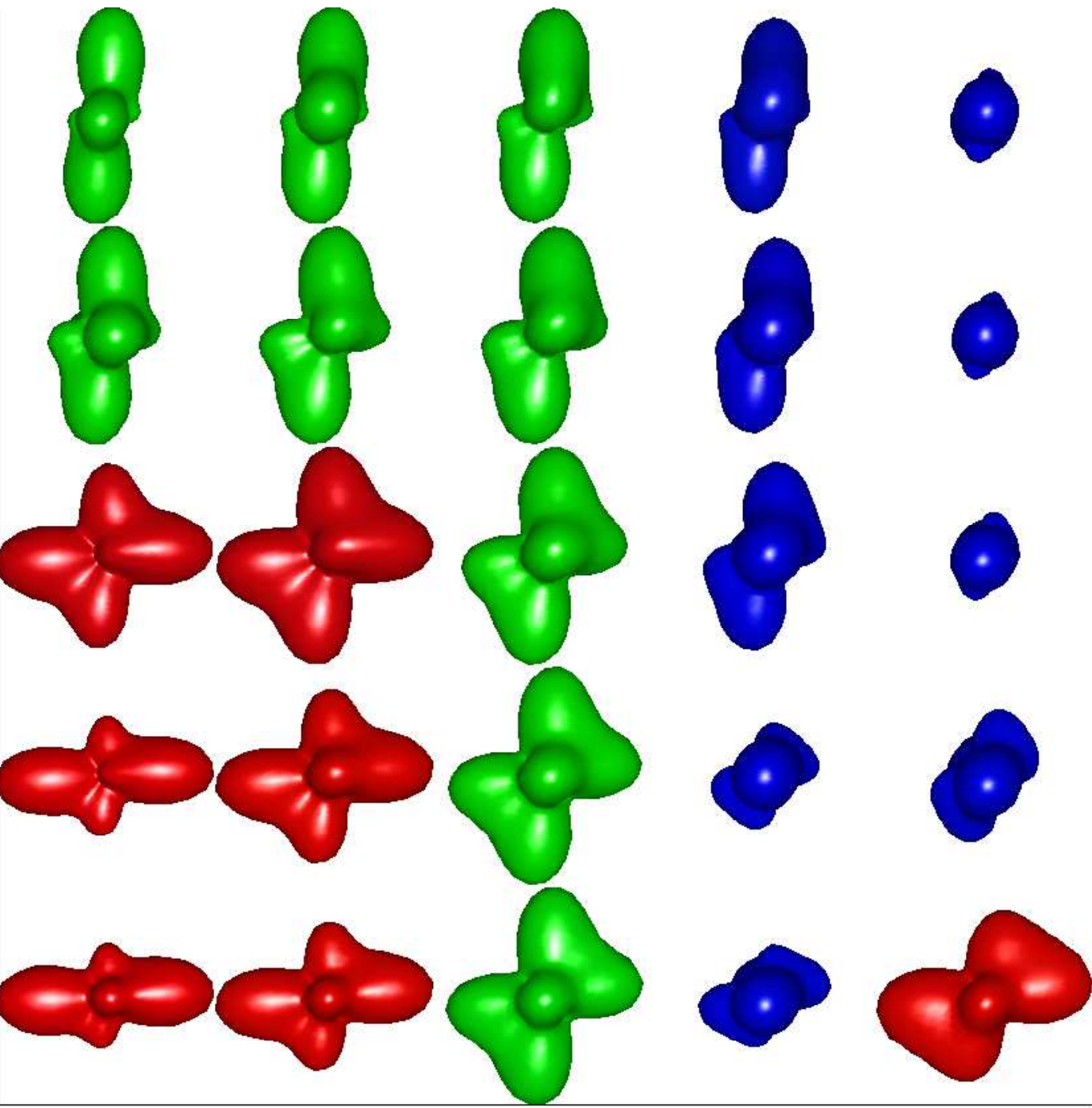}}\hspace{0.4cm}
	\subfigure[]{\includegraphics[scale=0.3]{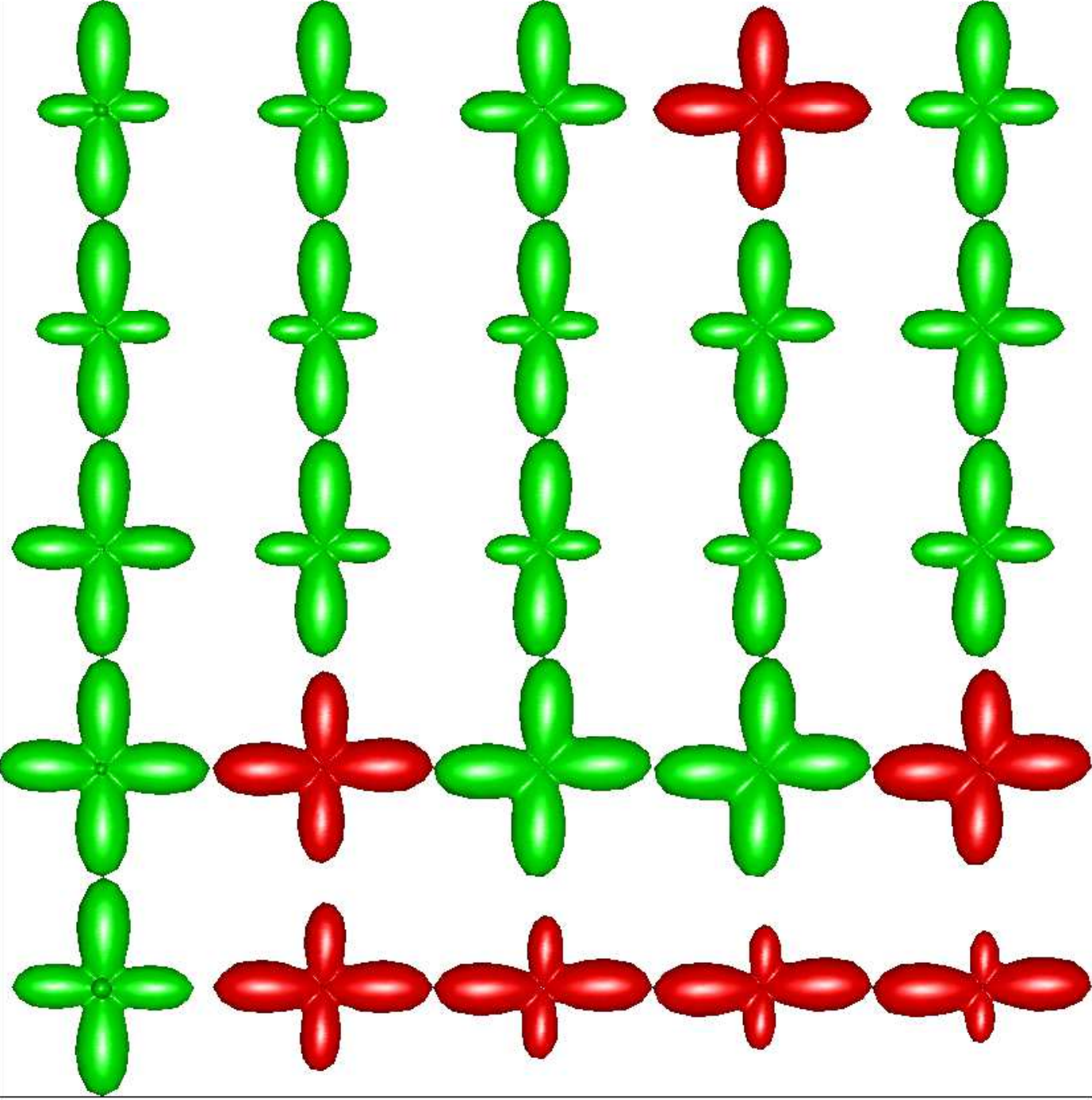}}	
	\caption{Examples of HOT fields: (a) rank-2,  (b) rank-4, and (c) rank-6}
	\label{Fields}	
\end{figure}

From a whole dMRI study, it is possible to estimate a grid of interconnected and related HOTs called tensor field. A HOT has a discrete graphical representation defined by parametrized surfaces called glyphs \cite{Ozarslan2003}. Figure \ref{Fields} shows examples of HOT fields of rank-2,4 and 6.

\subsection{Tucker decomposition of a Tensor}

Consider $\mathcal{T}$ $\in$ $\mathbb{K}^{I_1\times I_2 ... \times I_l}$
and $\mathbf{A}^{(1)}$ $\in$ $\mathbb{K}^{J_1\times I_1}$, $\mathbf{A}^{(2)}$ $\in$ $\mathbb{K}^{J_2\times I_2}$ and $\mathbf{A}^{(l)}$ $\in$ $\mathbb{K}^{J_l\times I_l}$. Then the Tucker mode-1 product $\mathcal{T}\cdot_1 \mathbf{A}^{(1)}$, mode-2 product $\mathcal{T}\cdot_2 \mathbf{A}^{(2)}$ and mode-$l$ product $\mathcal{T}\cdot_l \mathbf{A}^{(l)}$ are defined by
\begin{align*}
	\left( \mathcal{T}\cdot_1 \mathbf{A}^{(1)}\right)_{j_1i_2...i_l}=\displaystyle \sum_{i_1=1}^{I_1}\mathcal{T}_{i_1i_2...i_l}A^{(1)}_{j_1i_1}, \ \forall j_1,i_2,...,i_l, \\
	\left( \mathcal{T}\cdot_2 \mathbf{A}^{(2)}\right)_{i_1j_2...i_l}=\displaystyle \sum_{i_2=1}^{I_2}\mathcal{T}_{i_1i_2...i_l}A^{(2)}_{j_2i_2}, \ \forall i_1,j_2,...,i_l, \\
	\left( \mathcal{T}\cdot_l \mathbf{A}^{(l)}\right)_{i_1i_2...j_l}=\displaystyle \sum_{i_l=1}^{I_l}\mathcal{T}_{i_1i_2...i_l}A^{(l)}_{j_li_l}, \ \forall i_1,i_2,...,j_l.
\end{align*}
$\mathbb{K}$ may refer to $\mathbb{R}$ (real field) or $\mathbb{C}$ (complex field). A Tucker decomposition of a tensor $\mathcal{T}$ $\in$ $\mathbb{K}^{n \times ... \times^l n}$ is a decomposition of $\mathcal{T}$ of the form \cite{Tucker1964}
\begin{equation}
	\mathcal{T}=\mathcal{D}\cdot_1 \mathbf{A}^{(1)}\cdot_2 \cdots \mathbf{A}^{(l-1)} \cdot_l \mathbf{A}^{(l)},
	\label{Tuck}
\end{equation}
in which $\mathcal{D}$  $\in$ $\mathbb{K}^{n \times ... \times^l n}$ is known as the core tensor, and $\mathbf{A}^{(1)}$,  $\mathbf{A}^{(2)}$,..., $\mathbf{A}^{(l)}$ are matrices with column unitary vectors. If the decomposed tensor is symmetric and positive definite, $\mathbf{A}^{(1)}=\mathbf{A}^{(2)}=\mathbf{A}^{(l)}$. For a $l-$order tensor, equation \eqref{Tuck} is rewritten as follows:
\begin{equation}
	\mathcal{T}=\mathcal{D}\cdot_1\mathbf{A}\cdot_2\cdots \mathbf{A}\cdot_l \mathbf{A}=\left[\mathbf{A}\otimes^l \mathbf{A}\right] \operatorname{vec} \mathcal{D}.
	\label{kTucker}
\end{equation}
The Tucker decomposition is useful for dimensionality reduction of large tensor datasets. The actual data analysis can then be carried out in a space of lower dimensions. The Tucker approximation is also important when one wishes to estimate signal subspaces from tensor data.

\subsection{Tucker Decomposition Process (TDP)}

Based on the formulation given in equation \eqref{kTucker}, and inspired by \cite{Wilson2011}, we propose non-parametric Bayesian model for higher order tensors that we call \textit{Tucker decomposition process} (TDP). The TDP is a probabilistic distribution over HOT indexed in an independent variable. In the context of dMRI, the independent variable refers to the spatial coordinates $\mathbf{z}=\left[x,y\right]^\top $. Let $\mathcal{T}(\boldz)$ be a random field of higher order tensors. We say that $\mathcal{T}(\boldz)$ follows a TDP according to
\begin{equation}
	\mathcal{T}(\boldz) \sim \mathcal{TDP}(\mathcal{D},\boldA(\mathbf{z}),l)=\mathcal{D}\cdot_1 \boldA(\mathbf{z})\cdot_2\cdots \boldA(\mathbf{z})\cdot_l \boldA(\mathbf{z})=\left[\boldA (\mathbf{z})\otimes^l \boldA(\mathbf{z})\right] \operatorname{vec} \mathcal{D},
	\label{GM2}
\end{equation} 
where $\mathcal{D}$ is a $l-$order symmetric core tensor, and $\mathbf{A}$ is a $n\times R$ matrix with column unitary vectors. The variable $n$ is the tensorial dimension ($n=3$ for dMRI), and $R$ is the degree of decomposition ($R\leq n$). In our probabilistic model, we assume that each element of $\boldA$ follows an independent Gaussian process (GP) indexed by $\mathbf{z}$. Also, we establish that unique elements of the core tensor are random variables sampled from a spherical multivariate Gaussian distribution. 
The number of unique elements of a tensor depends on its order $l$. For example if $l=4$, we have $N_l=15$ unique elements in a $4_{th}$ order tensor. Also, we set $n=3$, and $R=n=3$. The prior distributions over the parameters in the \textit{TDP} are given by $\boldA(\boldz)$ with elements $A_{ij}(\mathbf{z}) \sim \mathcal{GP}(\mathbf{0},k(\mathbf{z}_i,\mathbf{z}_j))$ for $i,j=1,2,3$; $\operatorname{vec} \mathcal{D} \sim \mathcal{N}(\mathbf{0},c^2\boldI)$, with $c^2$ the common variance for the elements in $\operatorname{vec} \mathcal{D}$. For the Gaussian process with use an exponentiated quadratic kernel with length-scale $\theta$. HOT Fields of figure \ref{Fields} are samples from the TDP (rank-2,4 and 6) with $\theta=0.1$. The core tensors $\mathcal{D}$ are HOT data estimated from real dMRI.




\subsection{Bayesian inference for the Tucker decomposition process}

Given a finite set of higher order tensors $\mathcal{X}(\mathbf{Z})=\{\mathcal{D}(\boldz_i)\}_{i=1}^N$ obtained from solving the Stejskal-Tanner formula for different input locations, we use Bayesian inference to compute the posterior distribution for the HOT field,
\begin{align*}
	p(\mathcal{T}(\mathbf{z})|\mathcal{X}(\mathbf{Z})) \propto p(\mathcal{X}(\mathbf{Z})|\mathcal{T}(\mathbf{z})) p(\mathcal{T}(\mathbf{z})).
\end{align*}
We use the TDP as the prior for $p(\mathcal{T}(\mathbf{z}))$, and for the likelihood function, we assume each element from the HOT data 
follows an independent Gaussian distribution with the same variance $\sigma^2$. This leads to a likelihood with the form
\begin{equation*}
	p(\mathcal{X}(\mathbf{Z})|\mathcal{T}(\mathbf{z})) \propto \prod_{i=1}^{N}\exp\left(-\frac{1}{2\sigma^2} \lVert \mathcal{X}(\mathbf{z}_i)- \mathcal{T}(\mathbf{z}_i)\rVert_{F}^2 \right), 
\end{equation*}
where $\lVert \mathcal{A}-\mathcal{B} \rVert_{F}$ is the tensorial Frobenius distance of order $l$ given by
\begin{equation}
	\lVert \mathcal{A}-\mathcal{B}\rVert_F=\left(\displaystyle \sum_{i_1,...,i_l}^{3} \left( \mathcal{A}_{i_1,...,i_l}-\mathcal{B}_{i_1,...,i_l}\right)^2  \right)^{1/2} 
	\label{frob}
\end{equation}
Posterior distributions are also computed for matrix $\boldA(\boldz)$, the length-scale parameter $\theta$ (for which a log-normal prior is used), and the core tensor $\mathcal{D}$.
We use Markov chain Monte Carlo algorithms to sample in cycles. Metropolis-Hastings is used to sample the posterior for the length-scale $\theta$, and for the elements of the core tensor $\mathcal{D}$. To sample $\boldA(\boldz)$ we employ elliptical slice sampling \cite{Murray2010}.

\subsection{HOT prediction with the Tucker decomposition process}
Once we learn the posterior distribution for all the parameters, we compute the predictive distribution for
$p(\mathcal{T}(\mathbf{z}_*))$ in a new spatial position
$\mathbf{z}_*=[x_*,y_*]^{\top}$. We vectorize all the elements of $\boldA(\mathbf{z})$, and $\boldA(\mathbf{z}_*)$, in vectors $\mathbf{u}$ and $\mathbf{u}_*$, respectively. The predictive distribution $p\left( \mathbf{u}_*|\mathbf{u}\right)$ is again Gaussian. 
From the mean value for $\mathbf{u}_*$ obtained from $p\left( \mathbf{u}_*|\mathbf{u}\right)$, we compute $\boldA(\mathbf{z}_*)$ using equation \eqref{GM2}, finding an estimate for $\mathcal{T}(\mathbf{z}_*)$.

\subsection{Experimental setup and datasets}
We test the Tucker decomposition process in HOT fields of rank $2$,$4$ and $6$. We obtain synthetic HOT fields from a random generative model, and we estimate HOT data from a real dMRI study using the method proposed in \cite{Barmpoutis2010}. The dMRI study was obtained from a human brain of a healthy subject on a General Electric Signa HDxt 3.0T MR scanner, 8-channel quadrature brain coil for reception, and $90$ gradient directions with a value for $b$ equal to 1000 $S/mm^2$. The study contains $128\times128\times33$  images in axial plane. As ground-truth or gold standard we use the original HOT data (synthetic and real), then we downsample the HOT fields by a factor of two. The downsampled fields are the training sets. After we train the TDP, we compute the predictive distribution for the HOT fields. For rank-2 data, we compare our approach with direct linear interpolation \cite{Pajevic2002}, log-Euclidean interpolation \cite{Arsigny2006}, and generalized Wishart processes (GWP) \cite{Vargas2015}. For rank-4 and 6, we compare against direct linear interpolation because the other two methodologies are valid only for rank-2 tensors. For a quantitative evaluation, we calculate an error metric based on the tensorial Frobenius distance (see eq. \eqref{frob}) between the interpolated field and the respective ground-truth.

\section{Experimental results}
In this section we show quantitative and qualitative results in two
different dMRI data: a synthetic dataset, and a real dMRI study
acquired from a human subject. For both datasets, we estimate HOT
fields of rank 2,4 and 6. 

\subsection{HOT fields interpolation in synthetic data}

\subsubsection*{Rank-2 Results}

\begin{figure}
	\centering
	\subfigure[]{\includegraphics[scale=0.243]{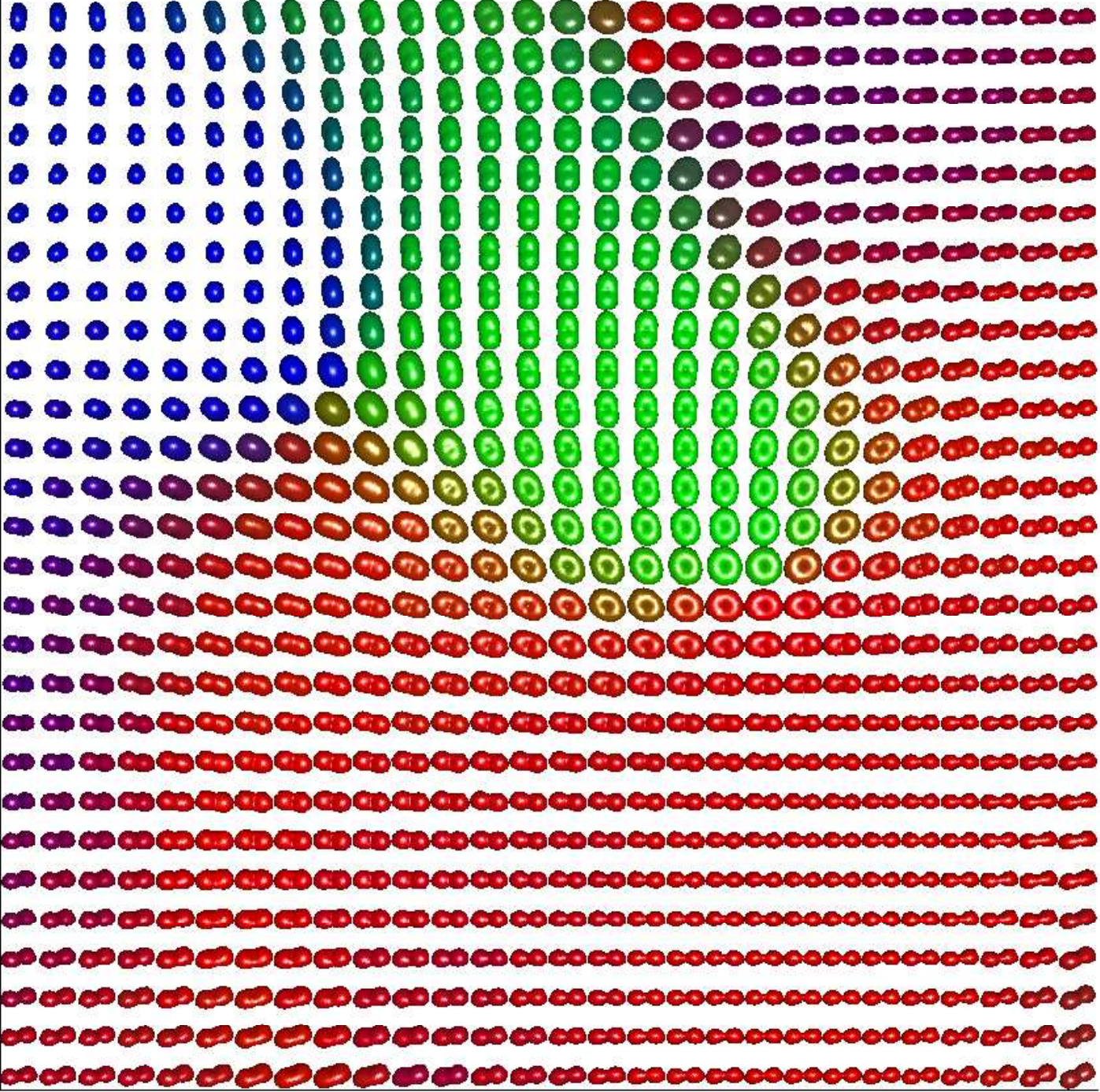}}
	\subfigure[]{\includegraphics[scale=0.145]{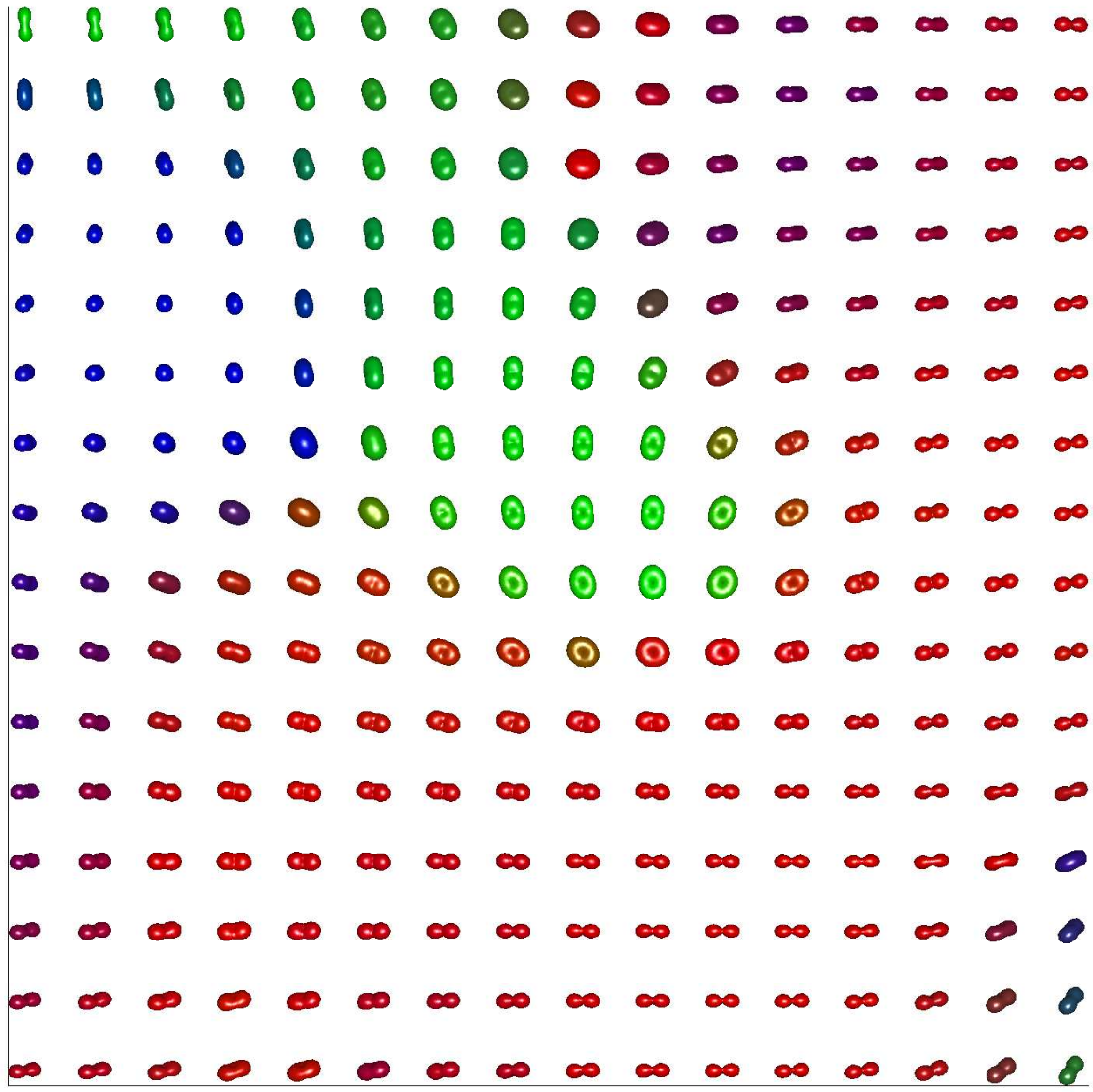}}
	\subfigure[]{\includegraphics[scale=0.243]{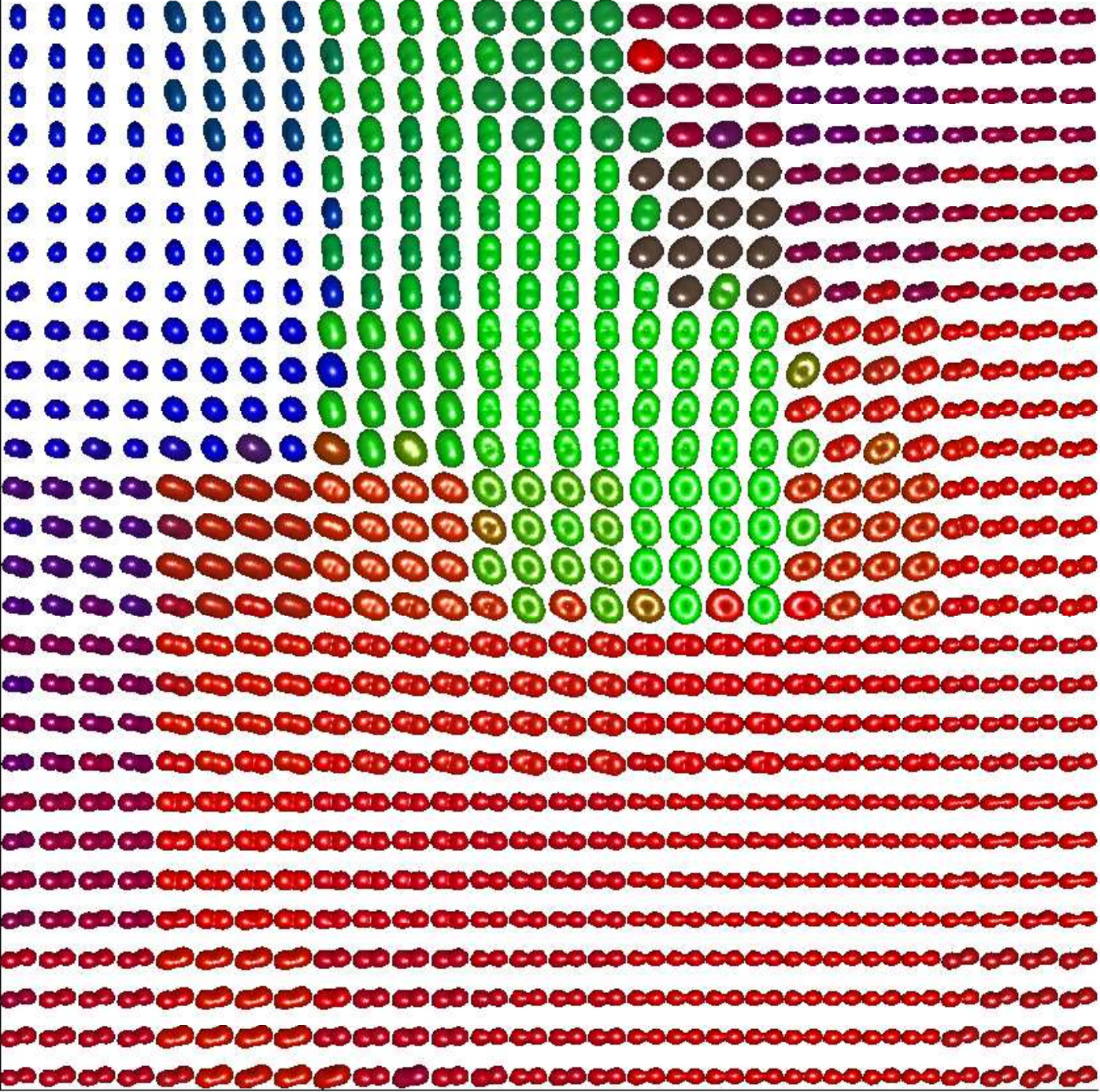}}
	\subfigure[]{\includegraphics[scale=0.243]{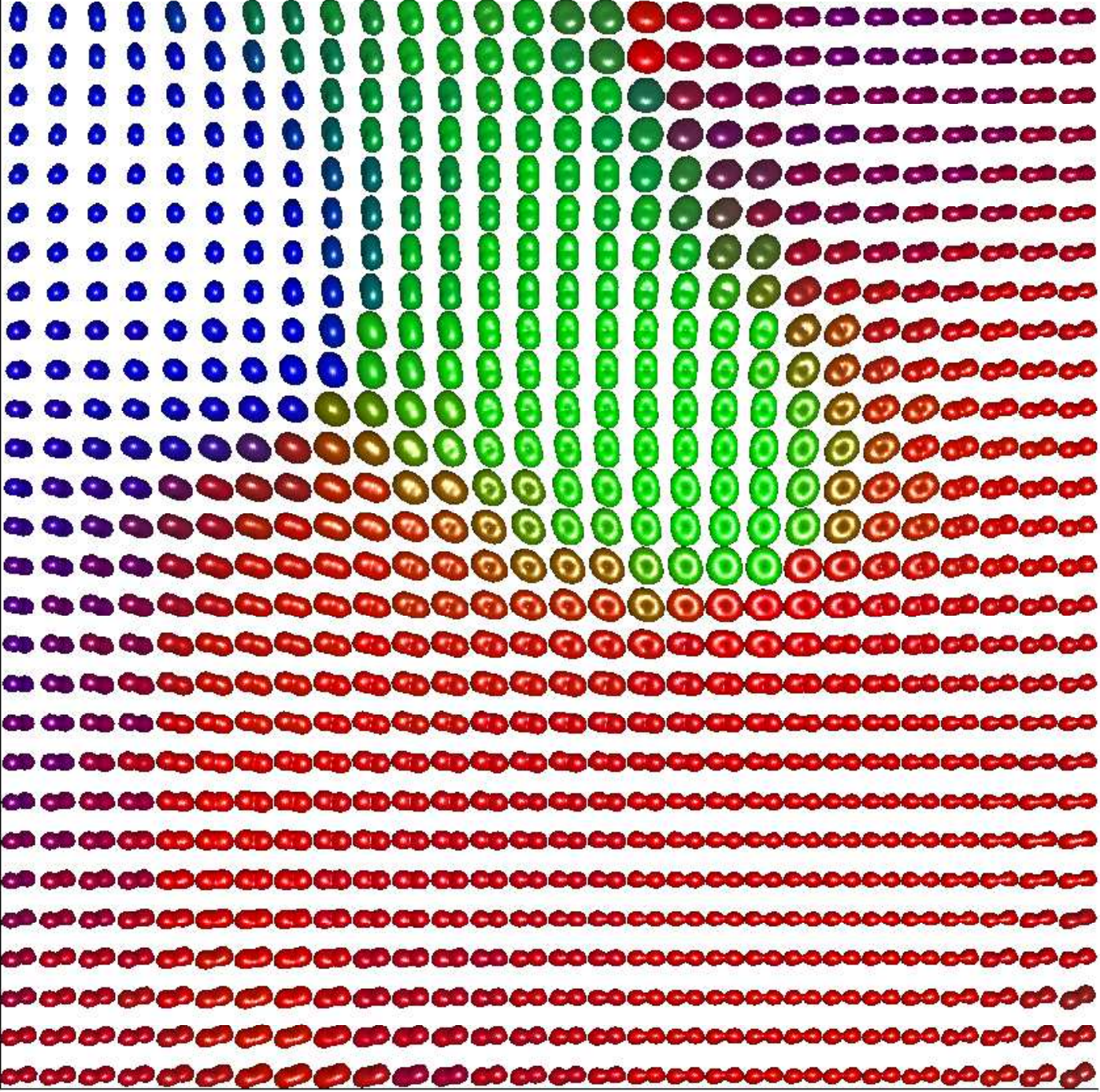}}
	\subfigure[]{\includegraphics[scale=0.243]{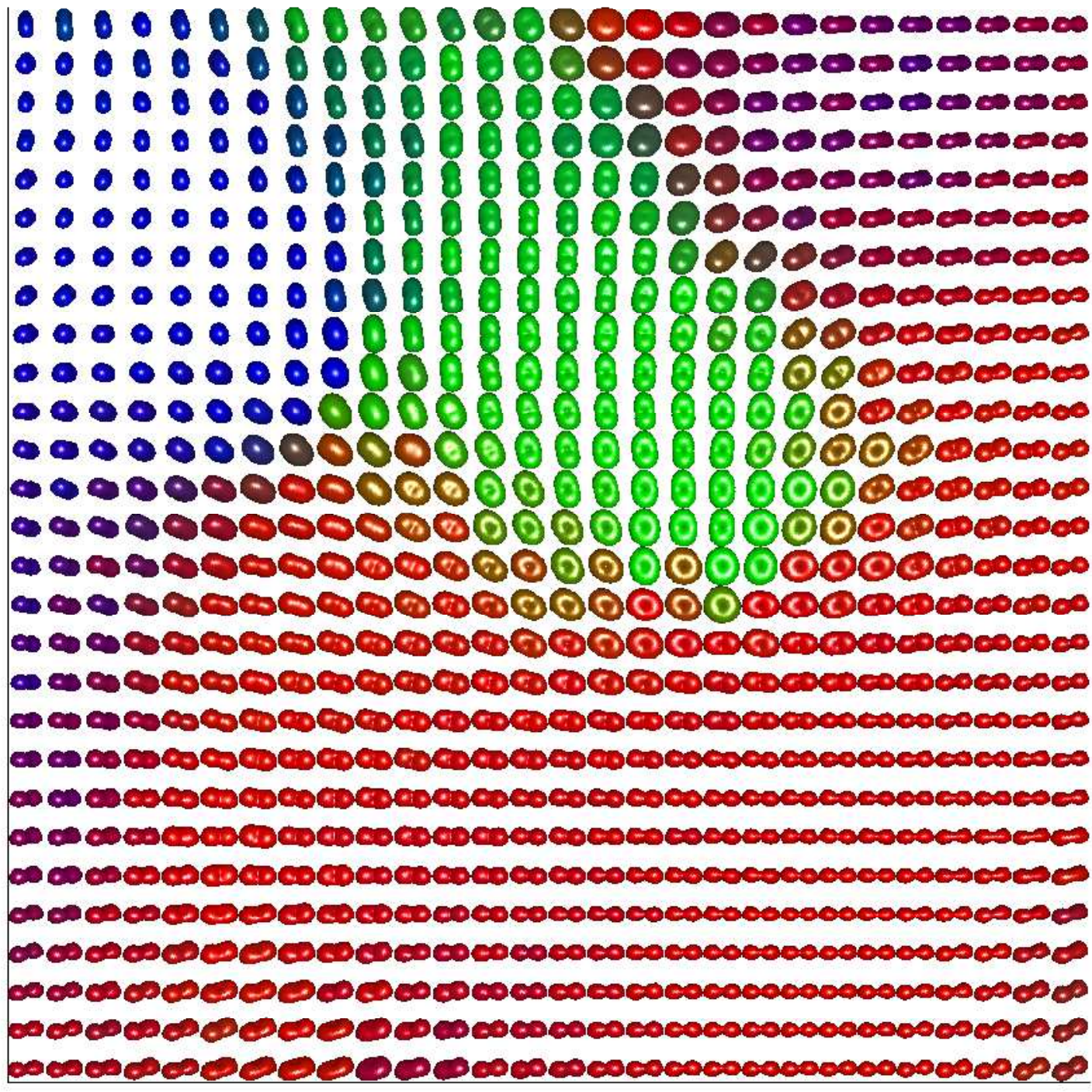}}
	\subfigure[]{\includegraphics[scale=0.243]{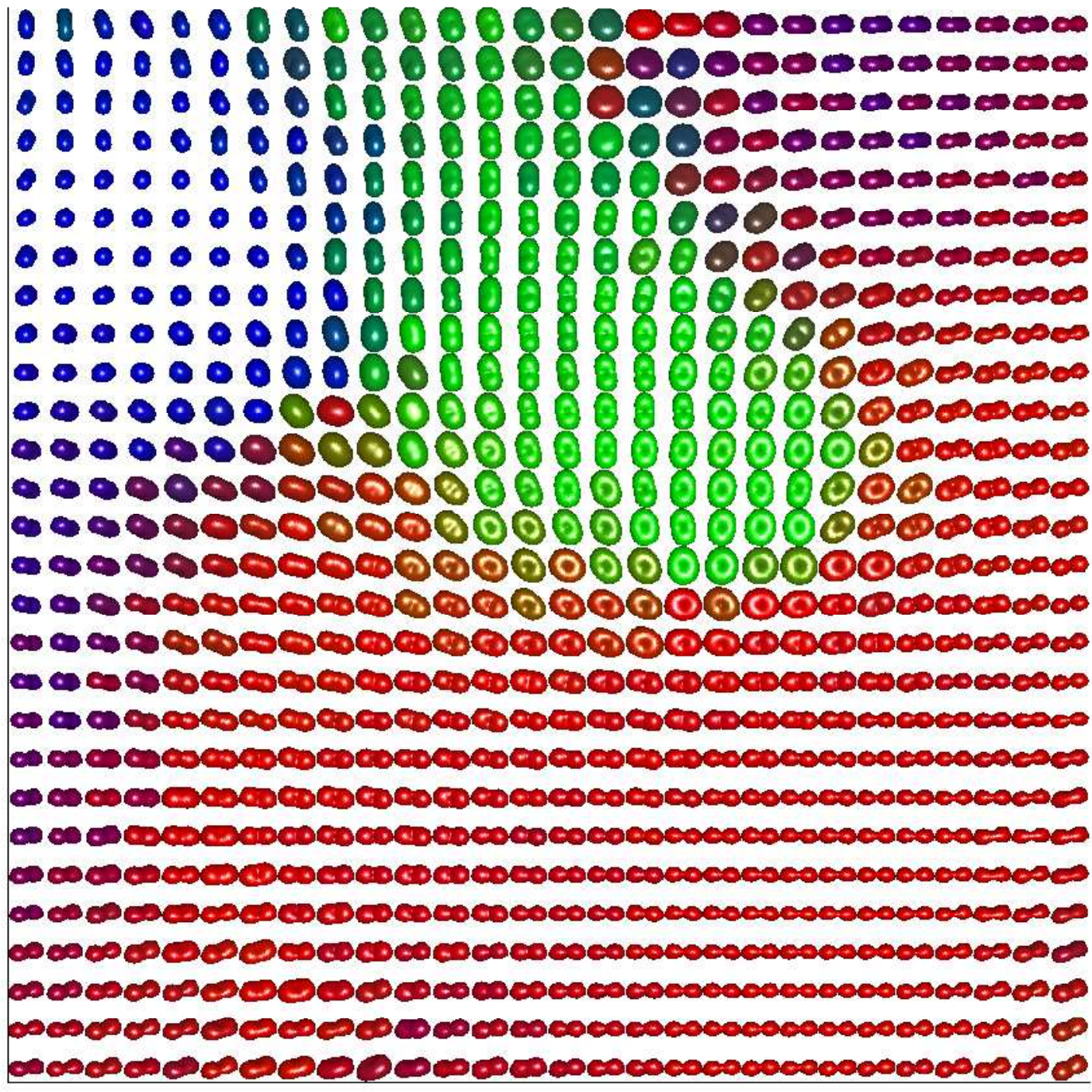}}	
	\caption{Graphic results for interpolation of a rank-2
		synthetic HOT field: (a) Ground-truth data,  (b) training
		data, (c) direct interpolation, (d) Tucker decomposition process, (e) generalized Wishart processes, and (f) log-Euclidean.}
	\label{st2}	
\end{figure}

Figure \ref{st2} and Table \ref{tab_ST2} show results for the rank-2
synthetic data. Note that probabilistic approaches (GWP and TDP)
exhibit better performance than direct interpolation and log-Euclidean
interpolation. One would expect GWP and TDP to exhibit a similar
performance. Although both methods have a similar construction, GWP only applies for rank-2 tensors. 

\begin{table}[ht!]
	\centering
	\caption{Frobenius distance for rank-2 synthetic HOT field}
	\begin{tabular}{cccc}
		\hline
		Direct interpolation & Log-Euclidean & Generalized Wishart
		processes & TDP  \\
		\hline
		$0.756\pm0.565$      & 	$0.432\pm 0.388$ & $0.279\pm0.211$ & $0.287\pm0.233$ \\
		\hline
	\end{tabular}%
	\label{tab_ST2}%
\end{table}%

\subsubsection*{Rank-4 and 6 results}
\begin{figure}
	\centering
	\subfigure[]{\includegraphics[scale=0.243]{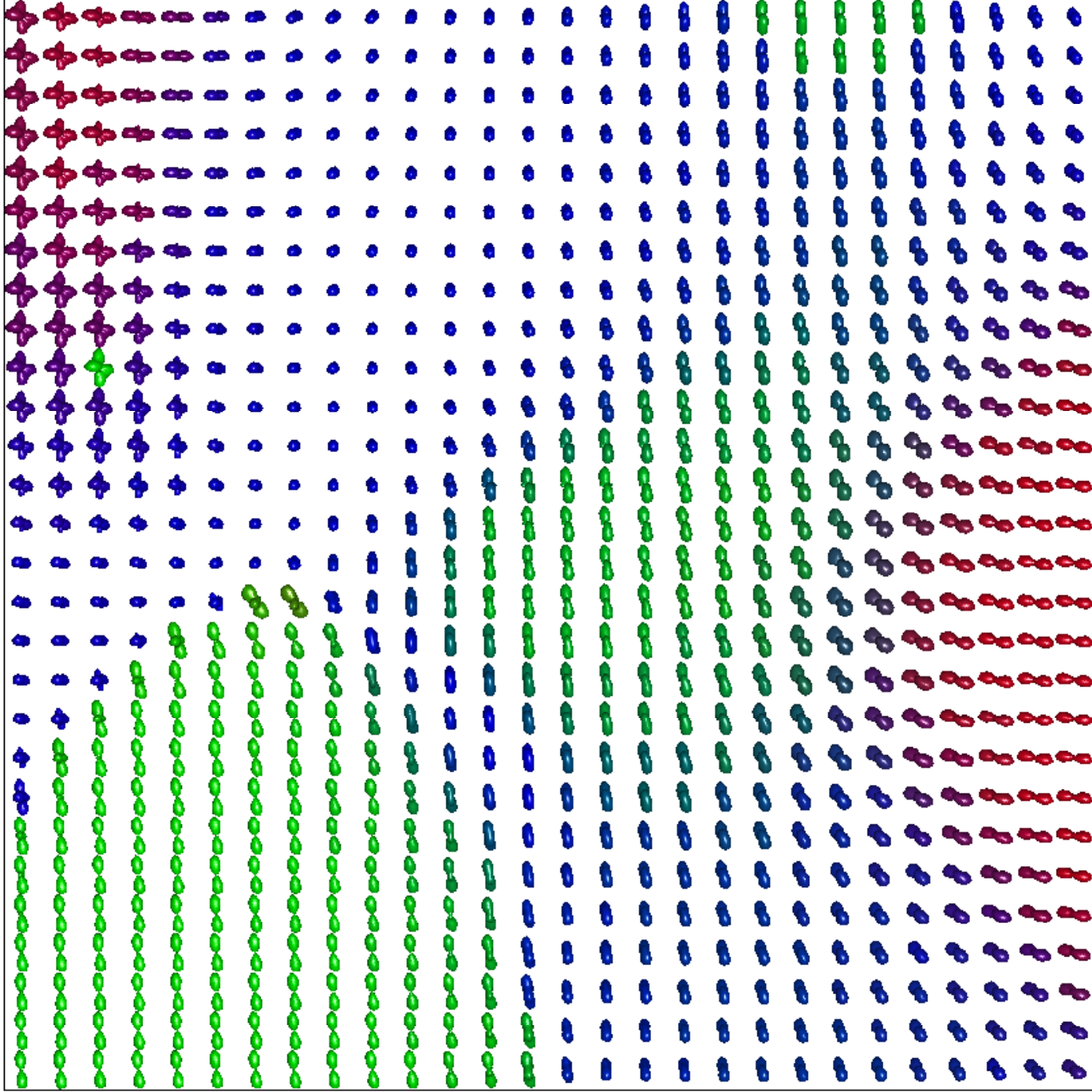}}
	\subfigure[]{\includegraphics[scale=0.145]{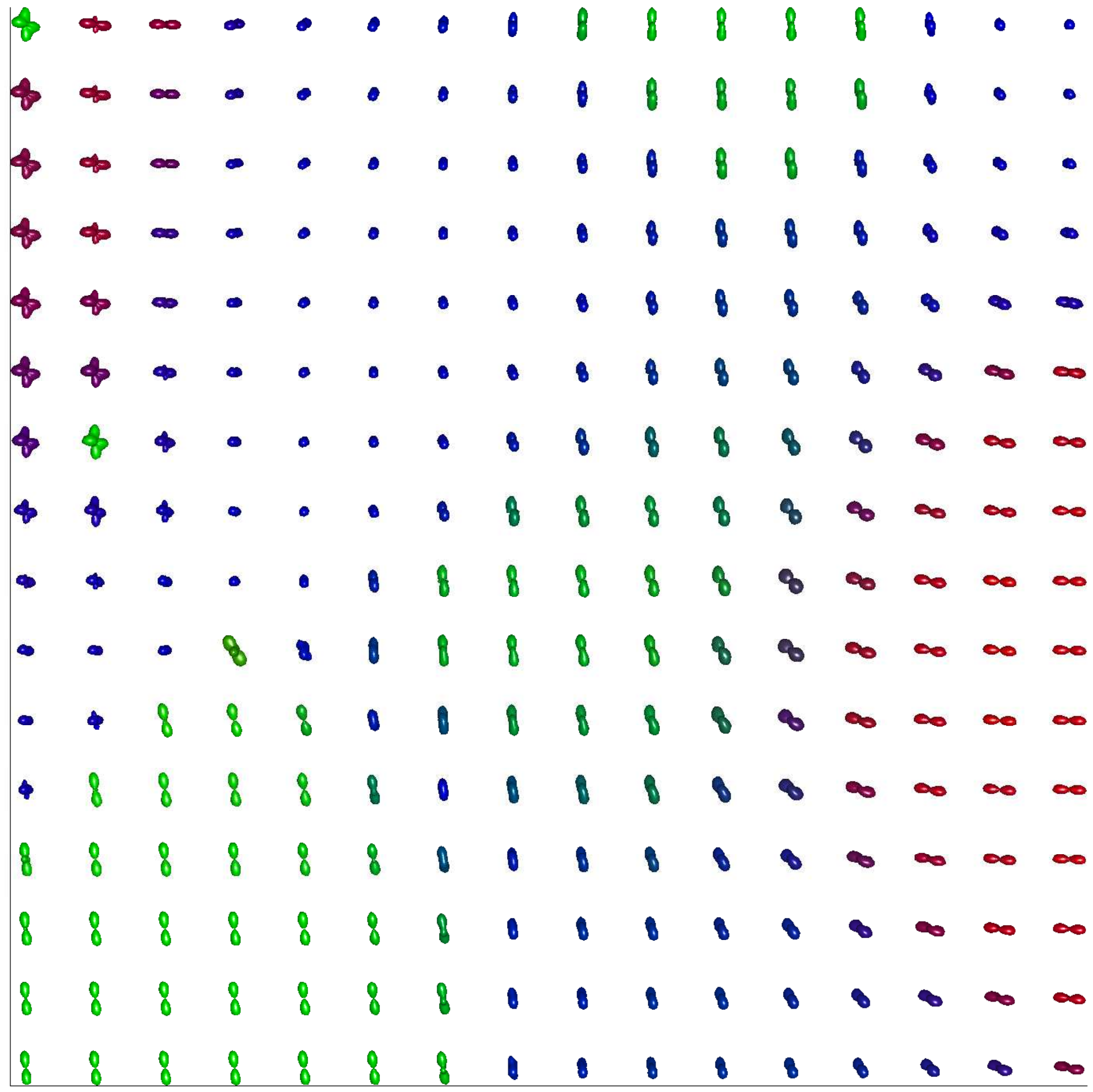}}
	\subfigure[]{\includegraphics[scale=0.243]{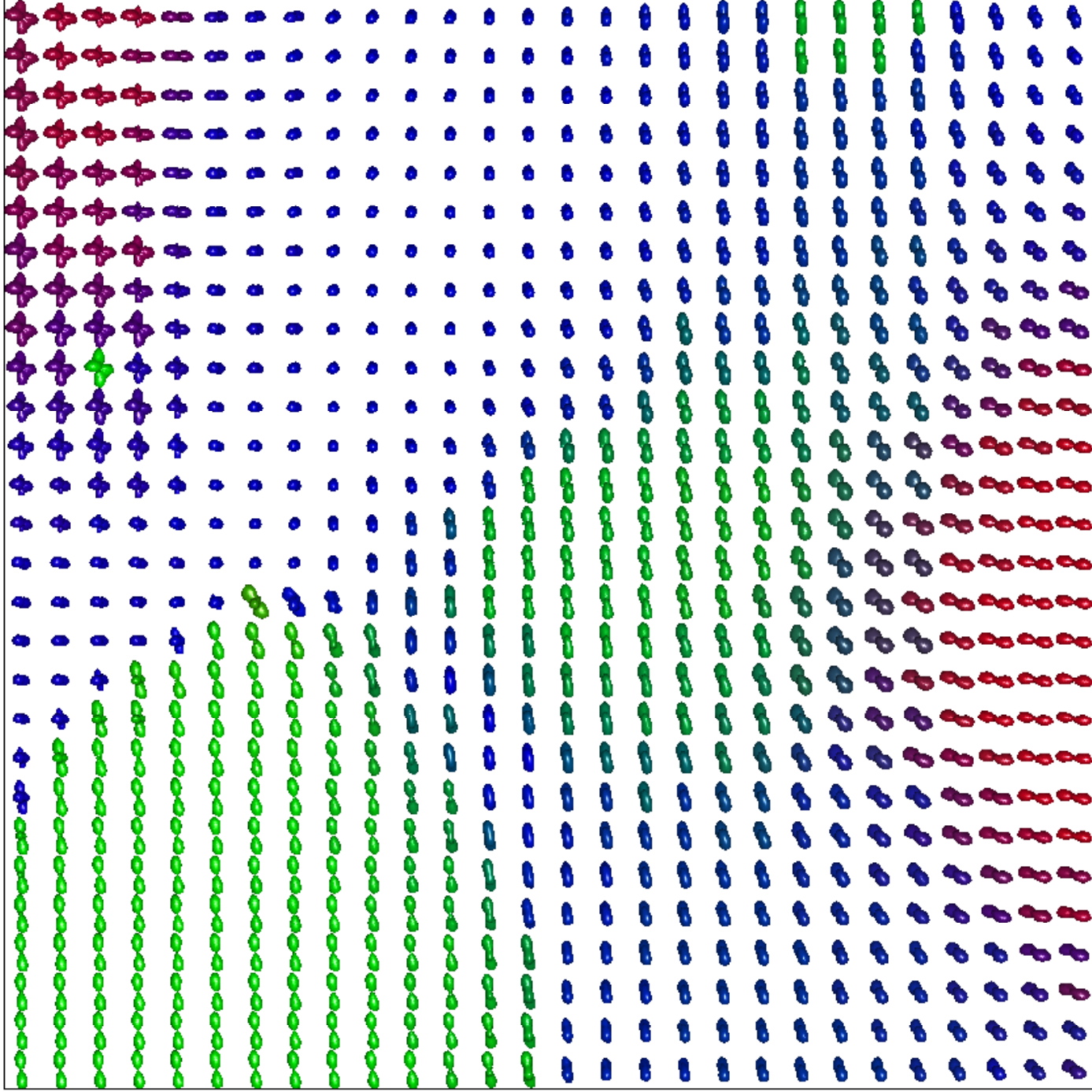}}
	\subfigure[]{\includegraphics[scale=0.243]{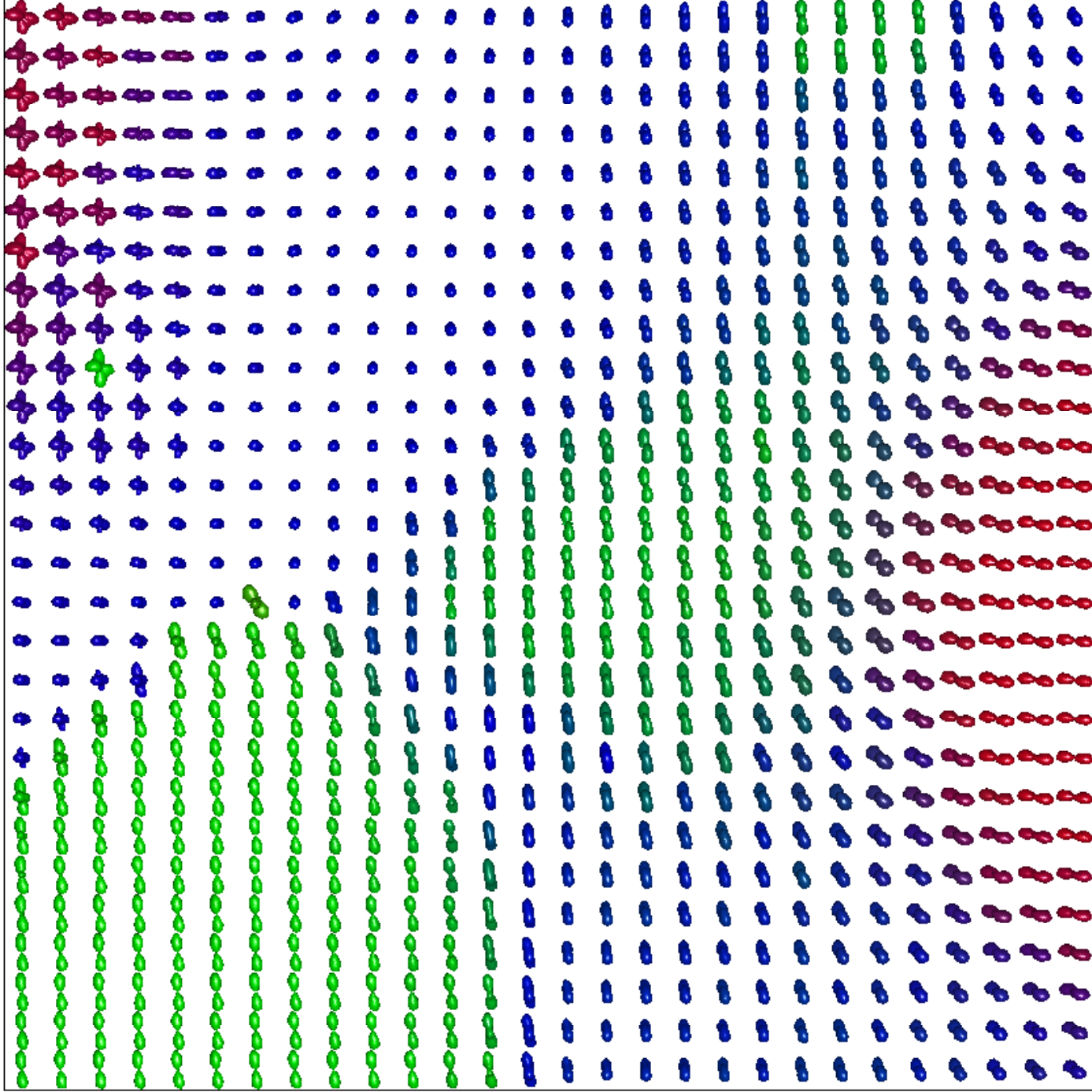}}	
	\caption{Graphic results for interpolation of rank-4 synthetic
		HOT field: (a) Ground-truth, (b) Training data, (c) linear
		interpolation, (d) Tucker decomposition process.}
	\label{st4}	
\end{figure}

\begin{figure}[ht!]
	\centering
	\subfigure[]{\includegraphics[scale=0.145]{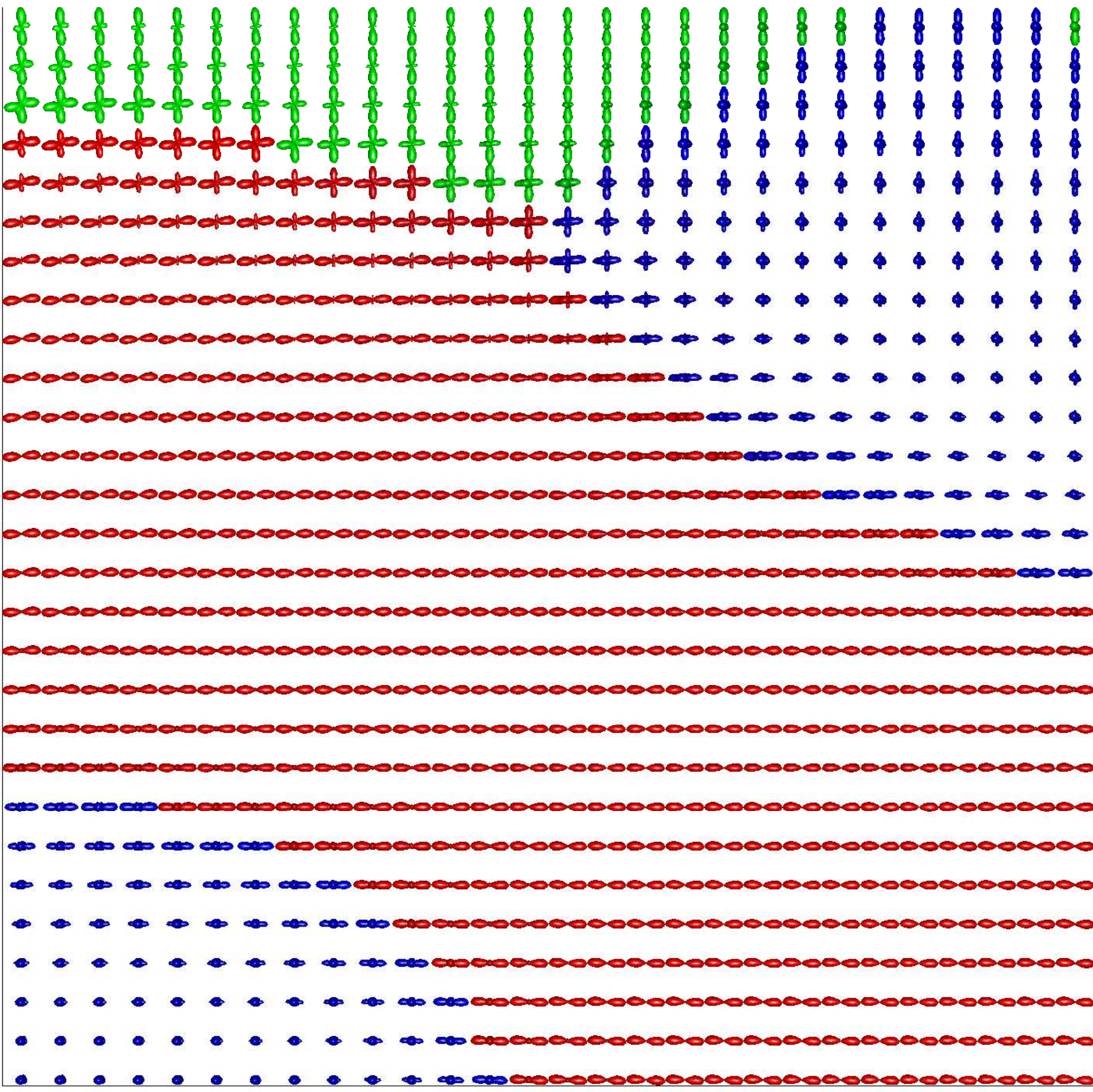}}
	\subfigure[]{\includegraphics[scale=0.145]{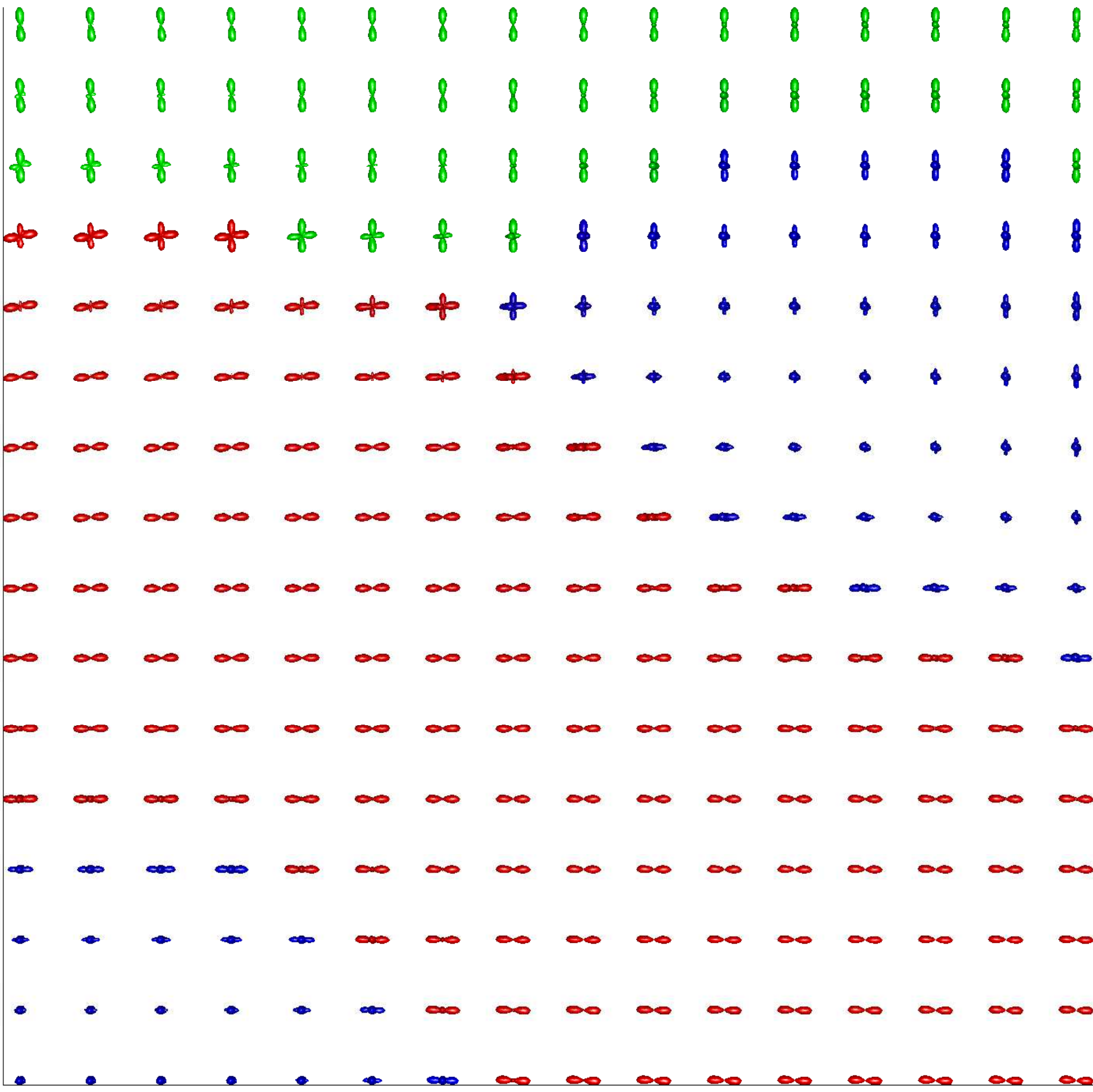}}
	\subfigure[]{\includegraphics[scale=0.145]{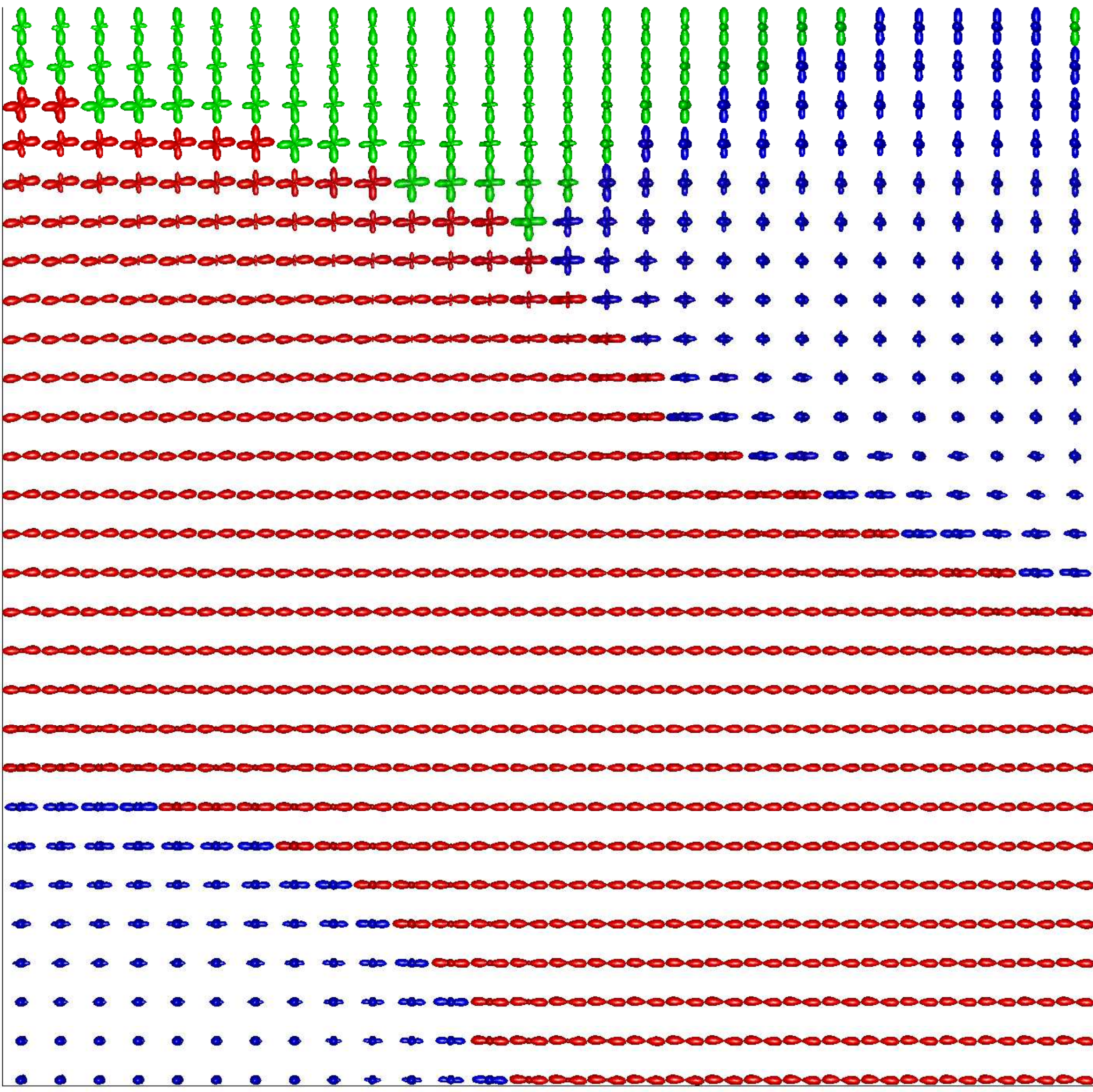}}
	\subfigure[]{\includegraphics[scale=0.145]{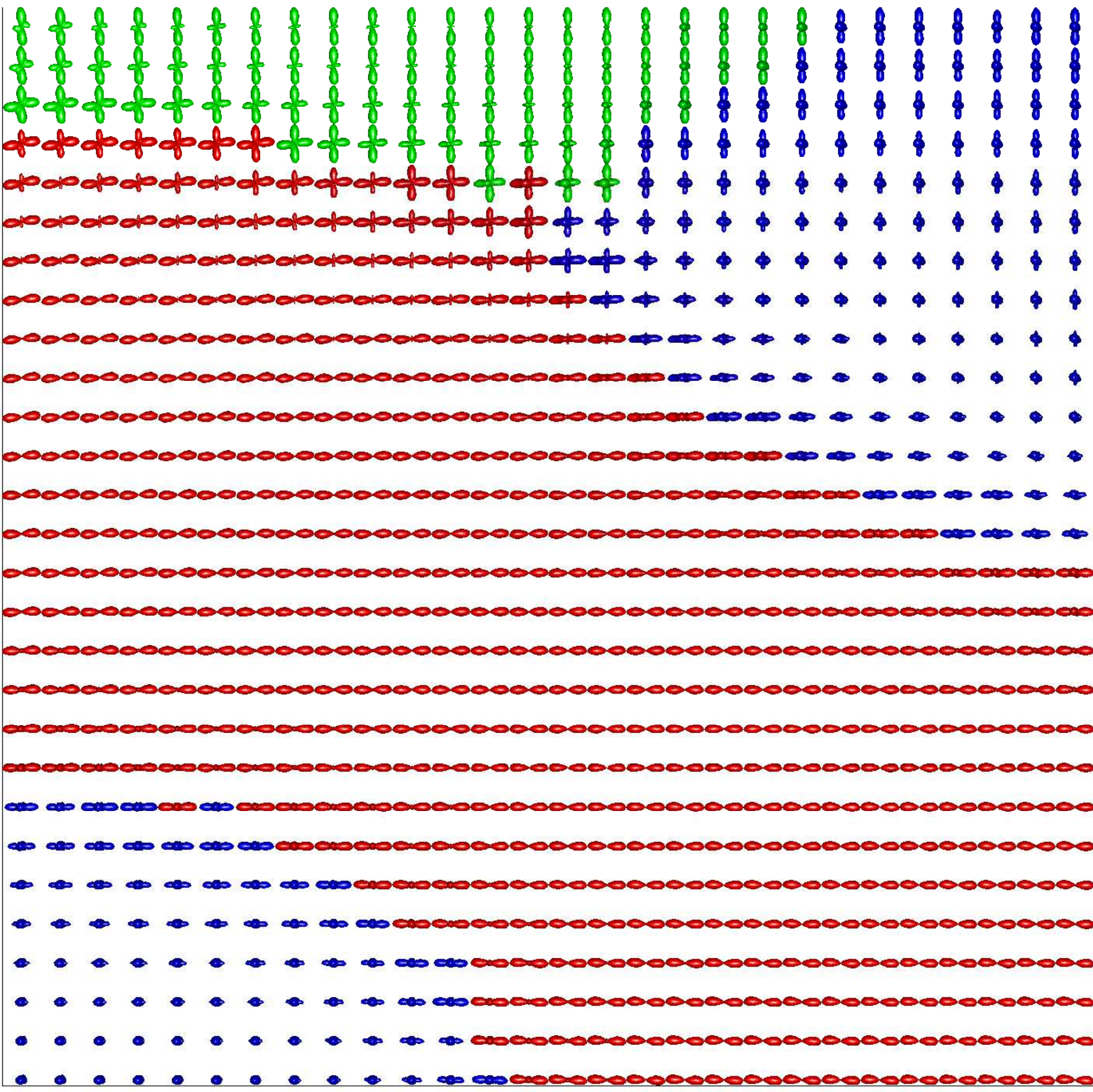}}	
	\caption{Graphic results for interpolation of rank-6 synthetic
		HOT field: (a) Ground-truth, (b) Training data, (c) linear
		interpolation, (d) Tucker decomposition process.}
	\label{st6}	
\end{figure}

Figures \ref{st4} and \ref{st6}, and Table \ref{tab1} show qualitative
and quantitative results for HOT interpolation in rank-4 and 6
synthetic data. As mentioned above, we only compare linear
interpolation against the TDP due to the lacking of methods for HOT interpolation.

\begin{table}[ht!]
	\centering
	\caption{Frobenius distance for rank-2 and 6 synthetic HOT fields}
	\begin{tabular}{ccc}
		\hline
		& Rank-4 & Rank-6 \\
		\hline
		TDP   & $0.840\pm0.968$ & $1.300\pm1.284$ \\
		Direct interpolation  & $1.386\pm1.457$ & $1.941\pm1.614$ \\
		\hline
	\end{tabular}%
	\label{tab1}%
\end{table}%

\subsection{HOT fields interpolation in real dMRI data} 

\subsubsection*{Rank-2 Results}

\begin{figure}[ht!]
	\centering
	\subfigure[]{\includegraphics[scale=0.244]{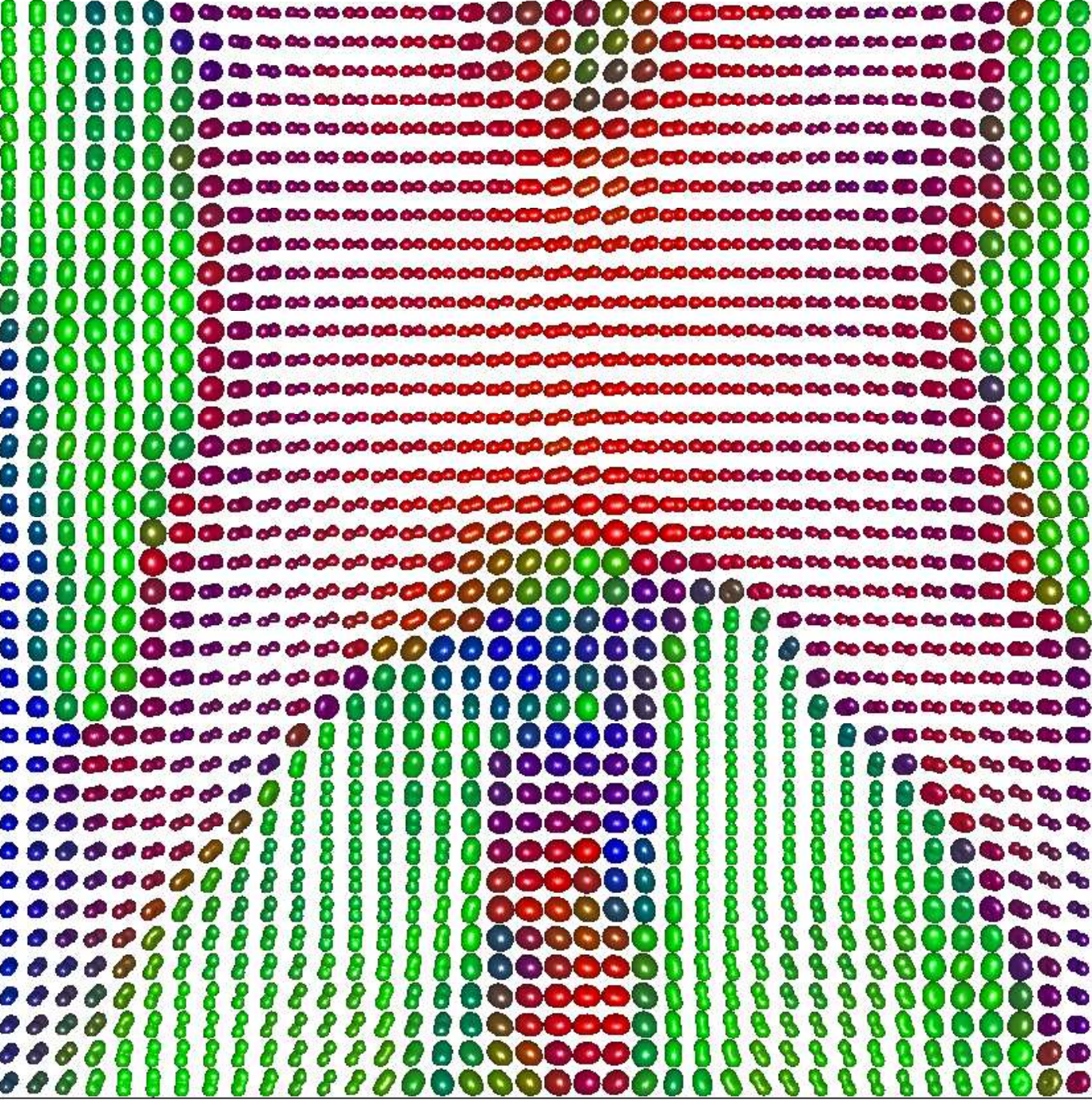}}
	\subfigure[]{\includegraphics[scale=0.145]{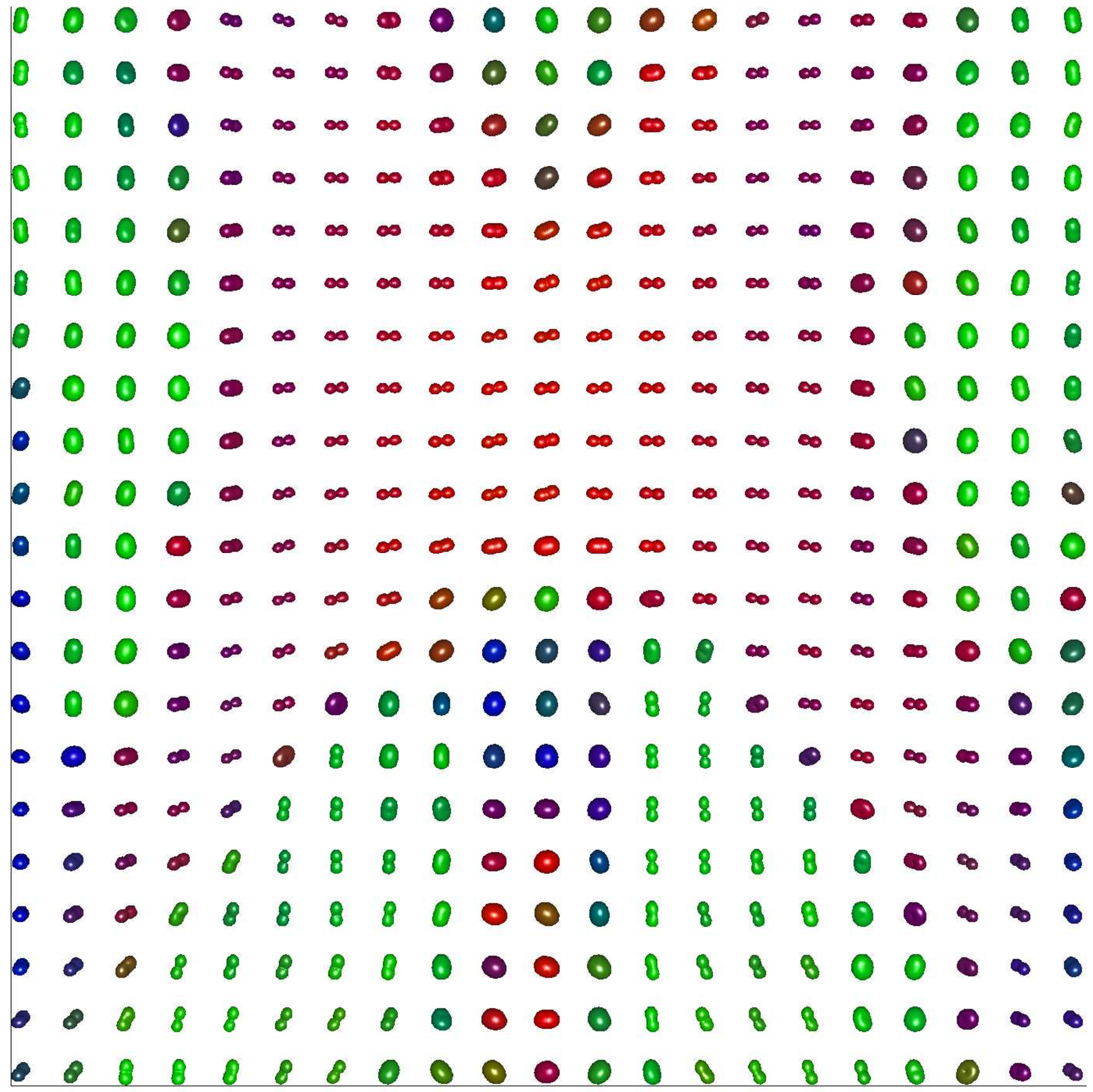}}
	\subfigure[]{\includegraphics[scale=0.244]{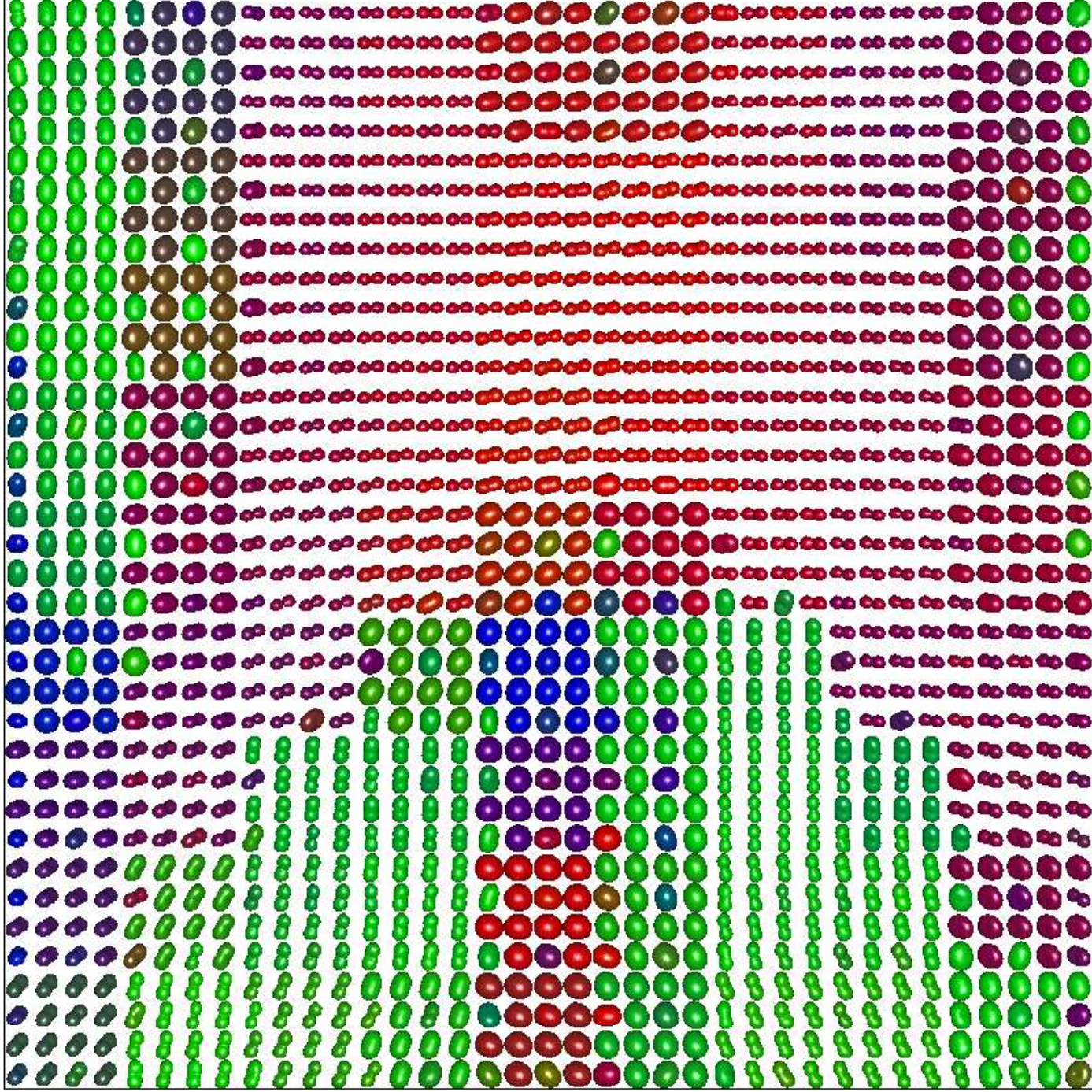}}
	\subfigure[]{\includegraphics[scale=0.244]{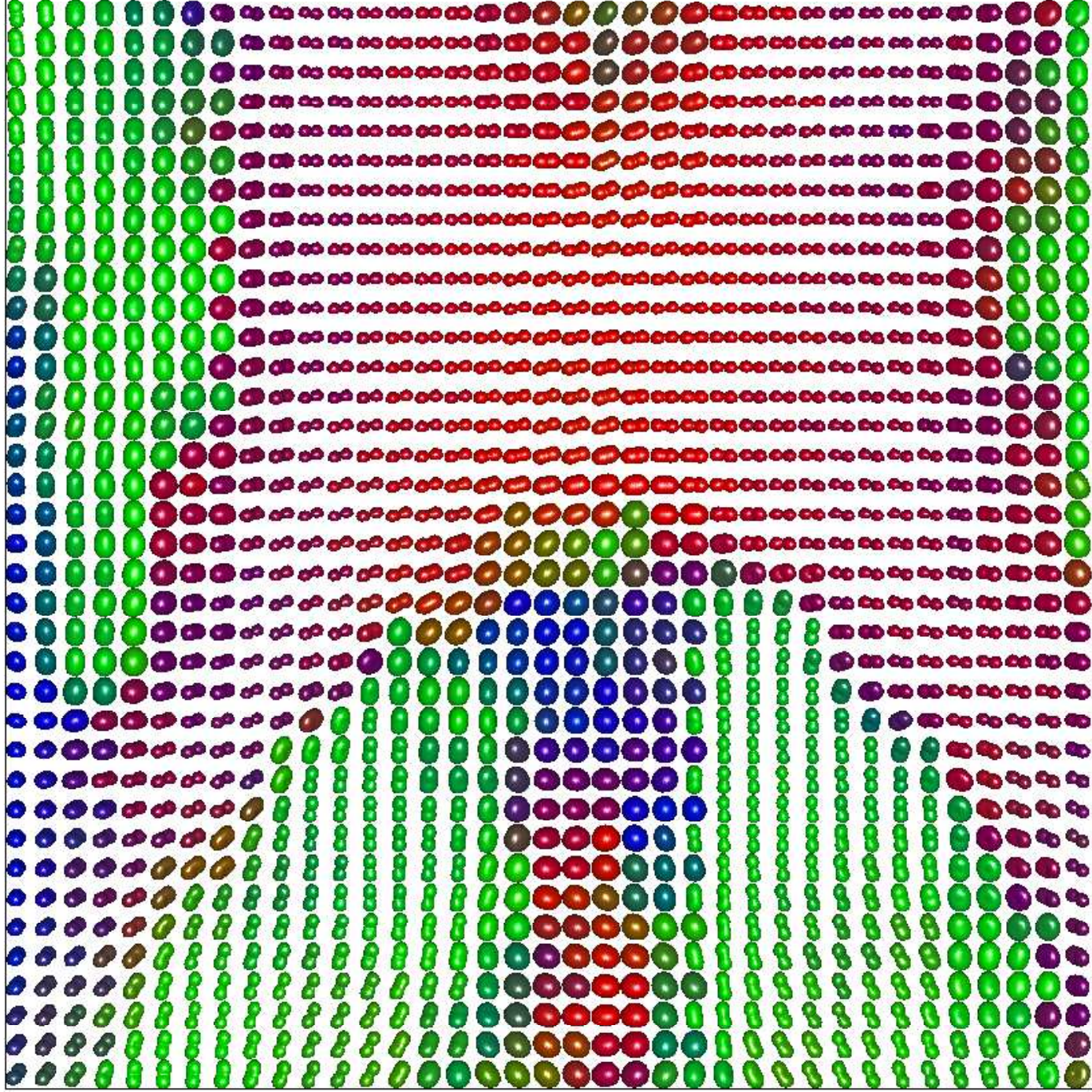}}
	\subfigure[]{\includegraphics[scale=0.244]{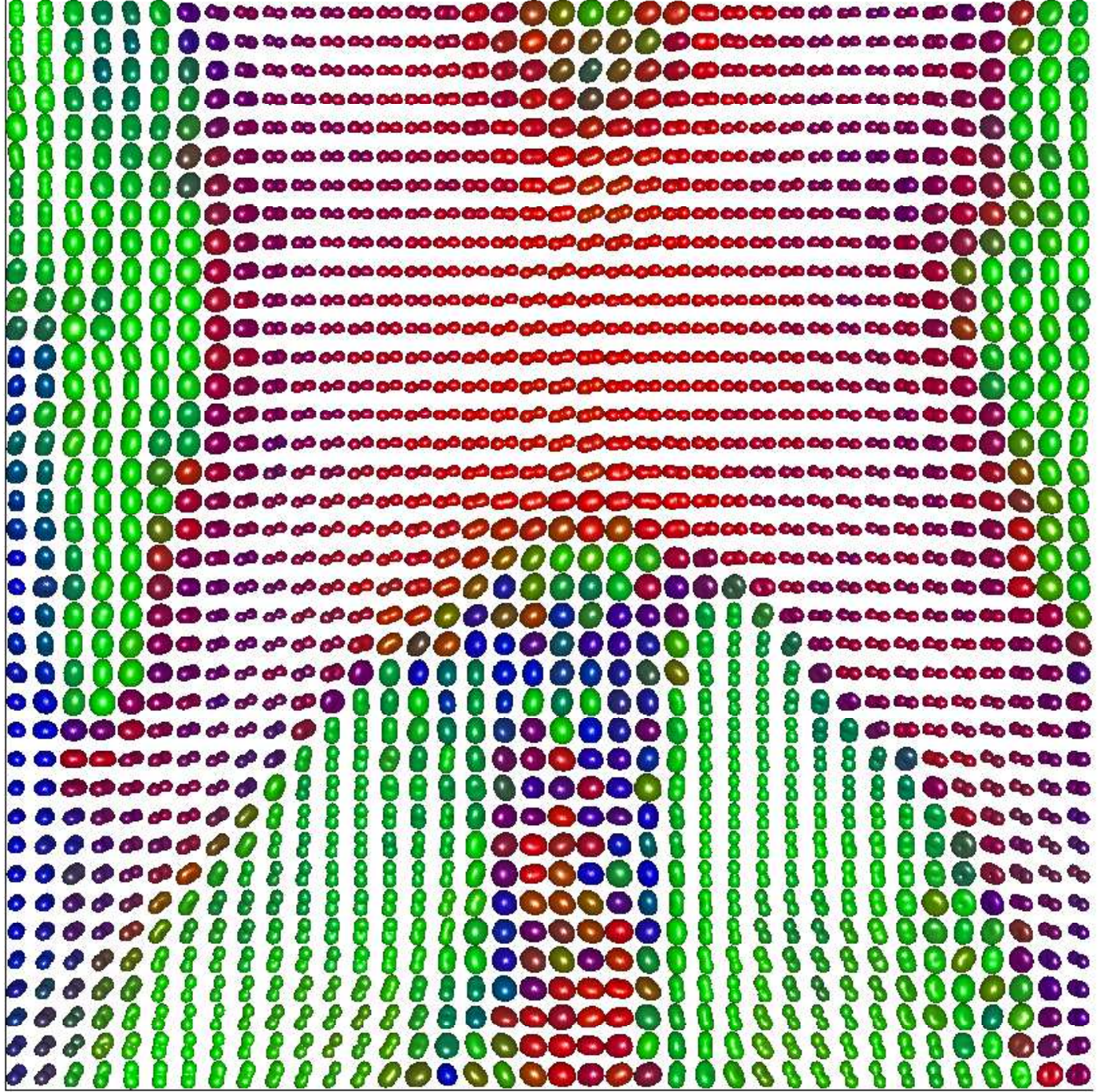}}
	\subfigure[]{\includegraphics[scale=0.244]{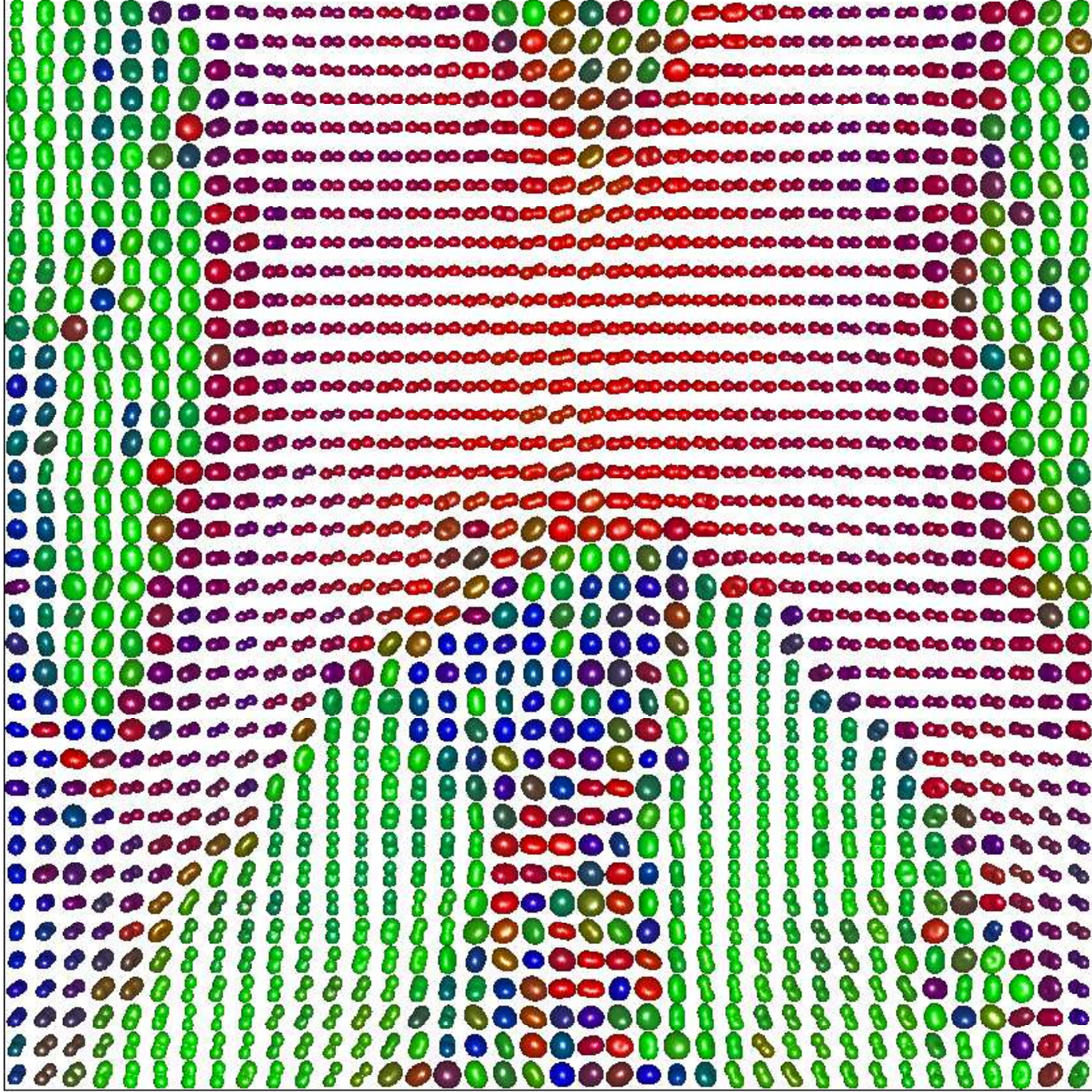}}	
	\caption{Graphic results for interpolation of rank-2 real HOT
		field: (a) Ground-truth data,  (b) training
		data, (c) direct interpolation, (d) Tucker decomposition process, (e) generalized Wishart processes, and (f) log-Euclidean.}
	\label{real2}
\end{figure}

Figure \ref{real2} and Table \ref{tab_R2} show results for the rank-2
real data. Similar to the results for the synthetic example, the
probabilistic methods offer better performance compared to the linear
and log-Euclidean interpolation.


\begin{table}[ht!]
	\centering
	\caption{Frobenius distance for rank-2 real HOT field}
	\begin{tabular}{cccc}
		\hline
		Direct interpolation & Log-Euclidean  & Generalized
		Wishart
		processes & Tucker process   \\
		\hline
		$0.506\pm0.463$ & $0.471\pm 0.367$ & $0.393\pm0.240$ & $0.388\pm0.294$ \\
		\hline
	\end{tabular}%
	\label{tab_R2}%
\end{table}%

\subsubsection*{Rank-4 and 6 results}
\begin{figure}[ht!]
	\centering
	\subfigure[]{\includegraphics[scale=0.243]{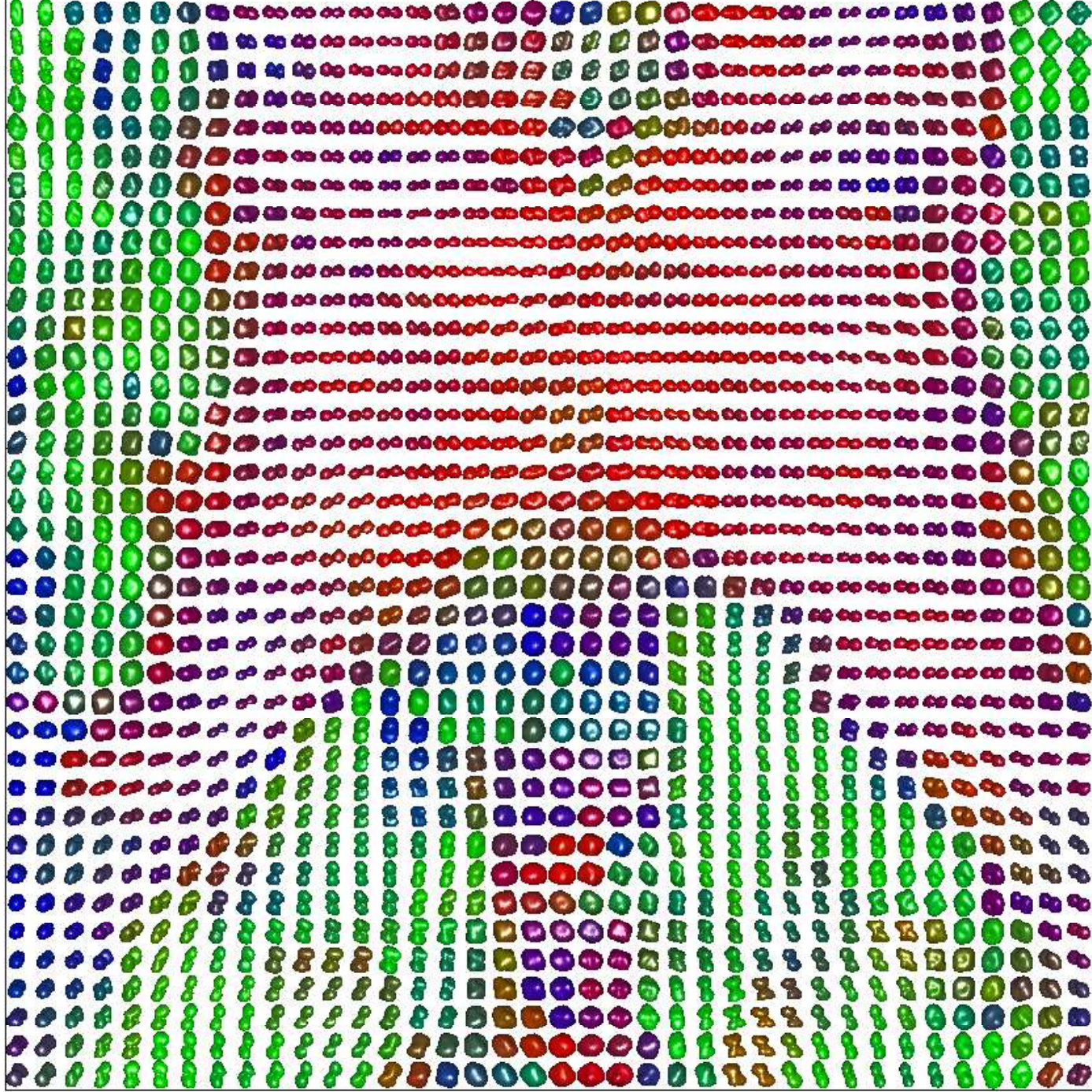}}
	\subfigure[]{\includegraphics[scale=0.145]{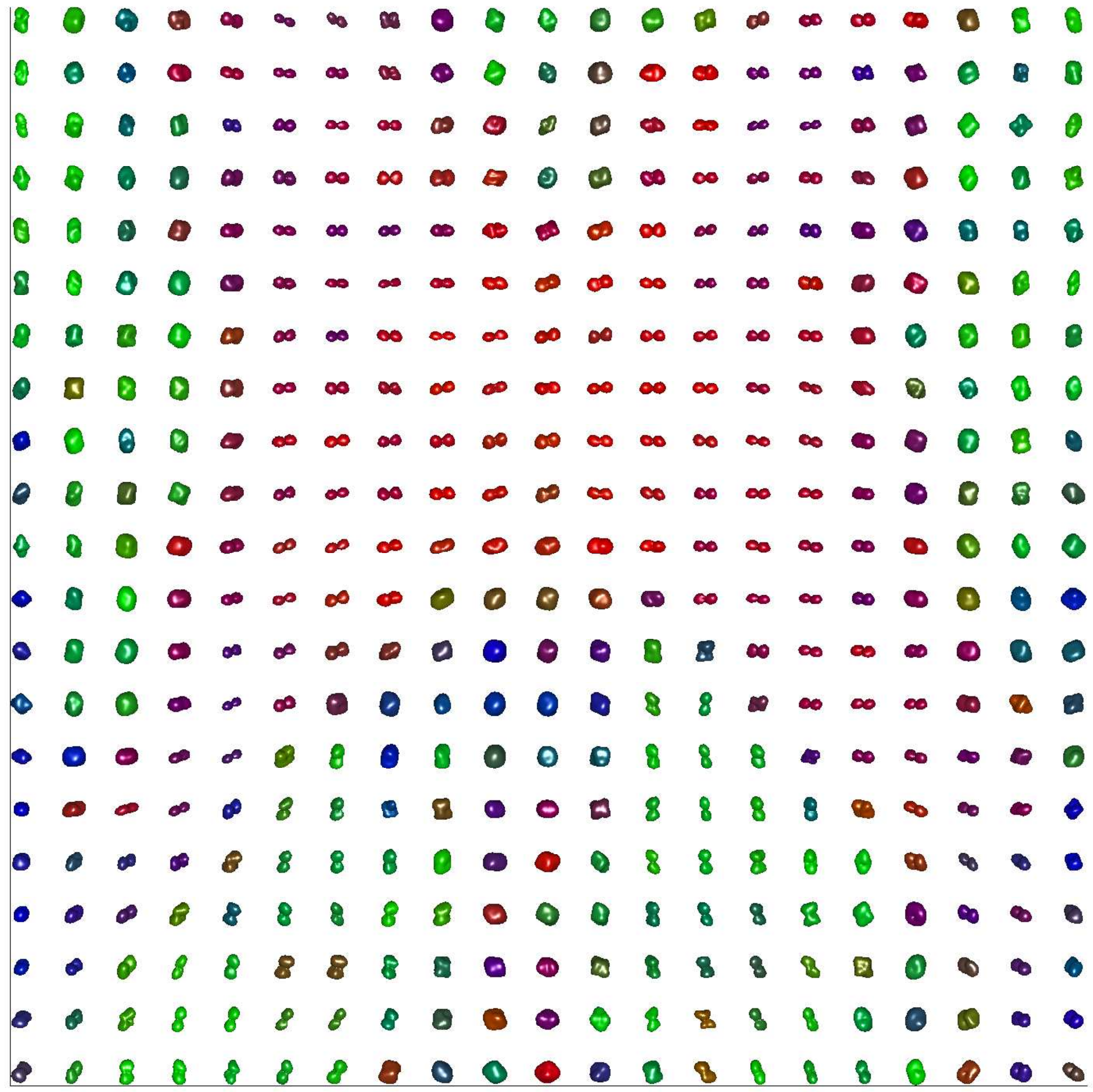}}
	\subfigure[]{\includegraphics[scale=0.243]{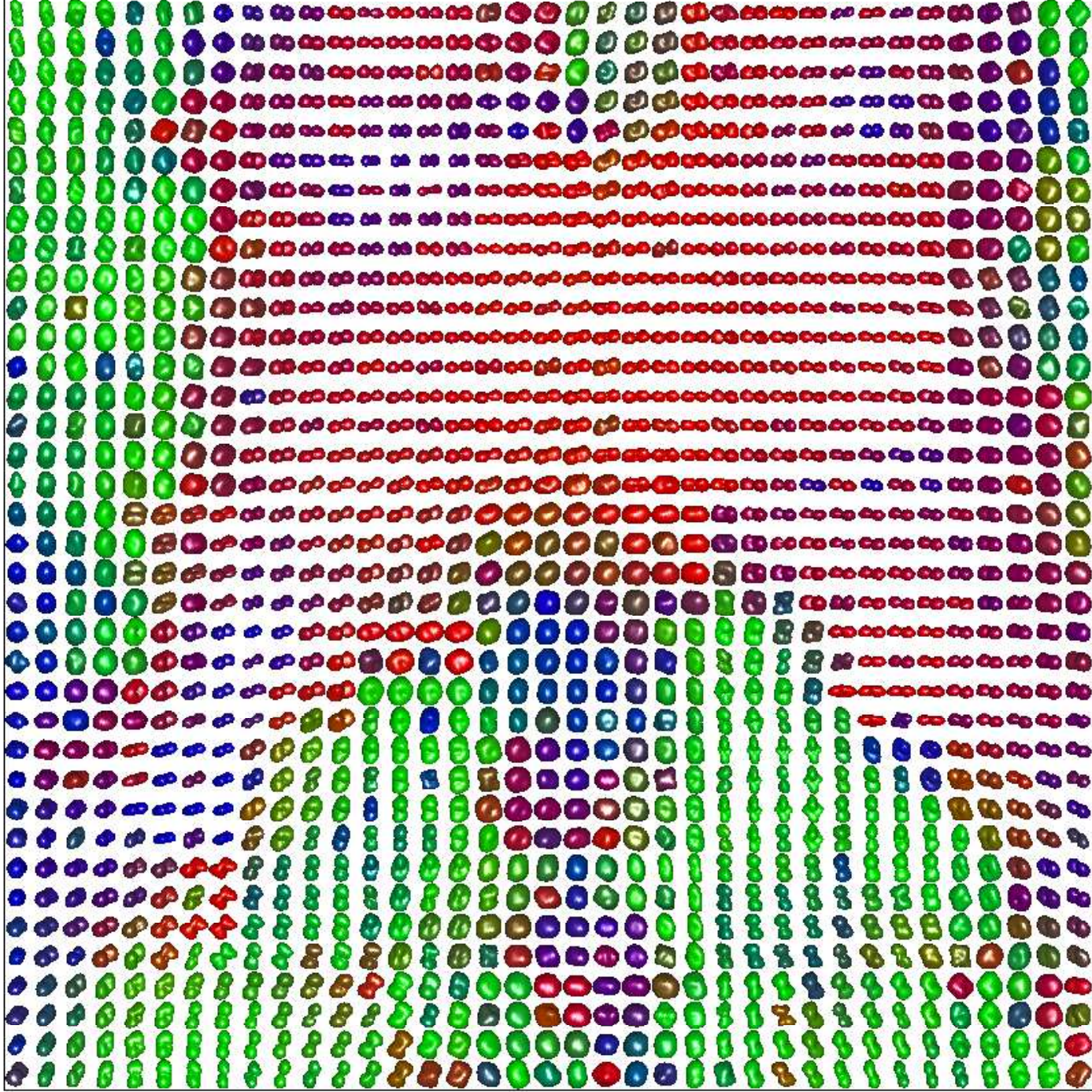}}
	\subfigure[]{\includegraphics[scale=0.243]{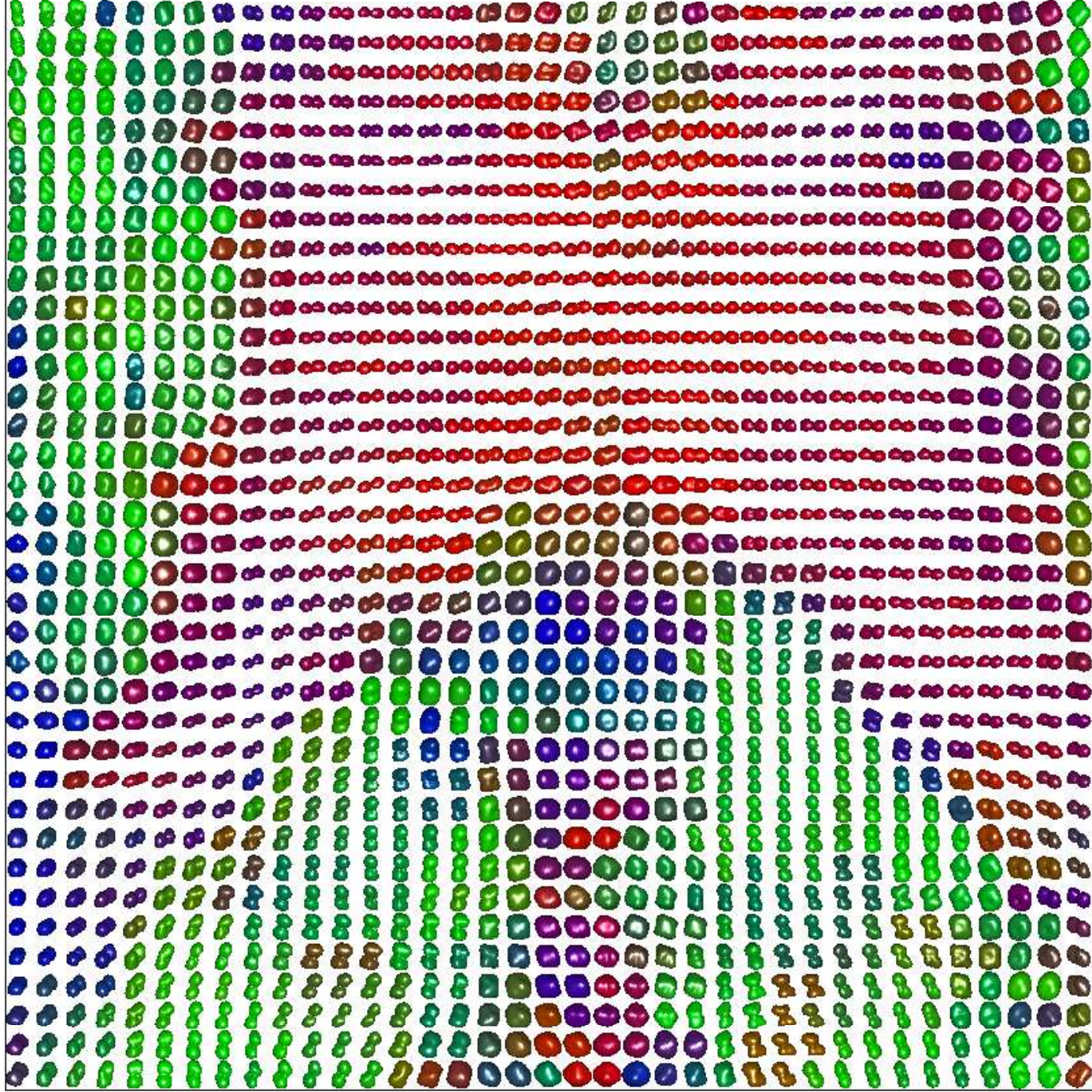}}	
	\caption{Graphic results for interpolation of rank-4 real HOT
		field: (a) Ground-truth, (b) Training data, (c) linear
		interpolation, (d) Tucker decomposition process.}
	\label{real4}	
\end{figure}

\begin{figure}[ht!]
	\centering
	\subfigure[]{\includegraphics[scale=0.145]{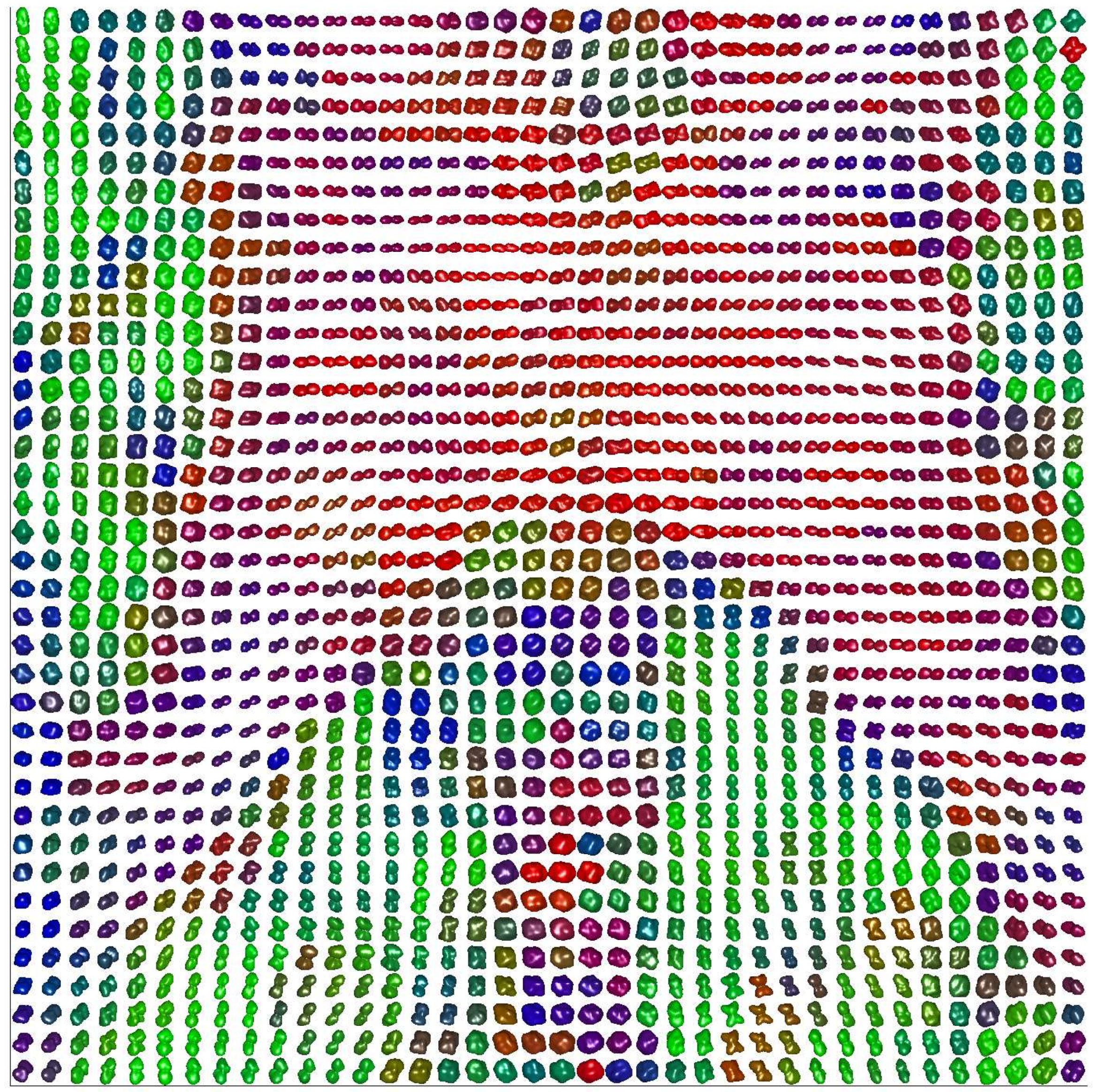}}
	\subfigure[]{\includegraphics[scale=0.145]{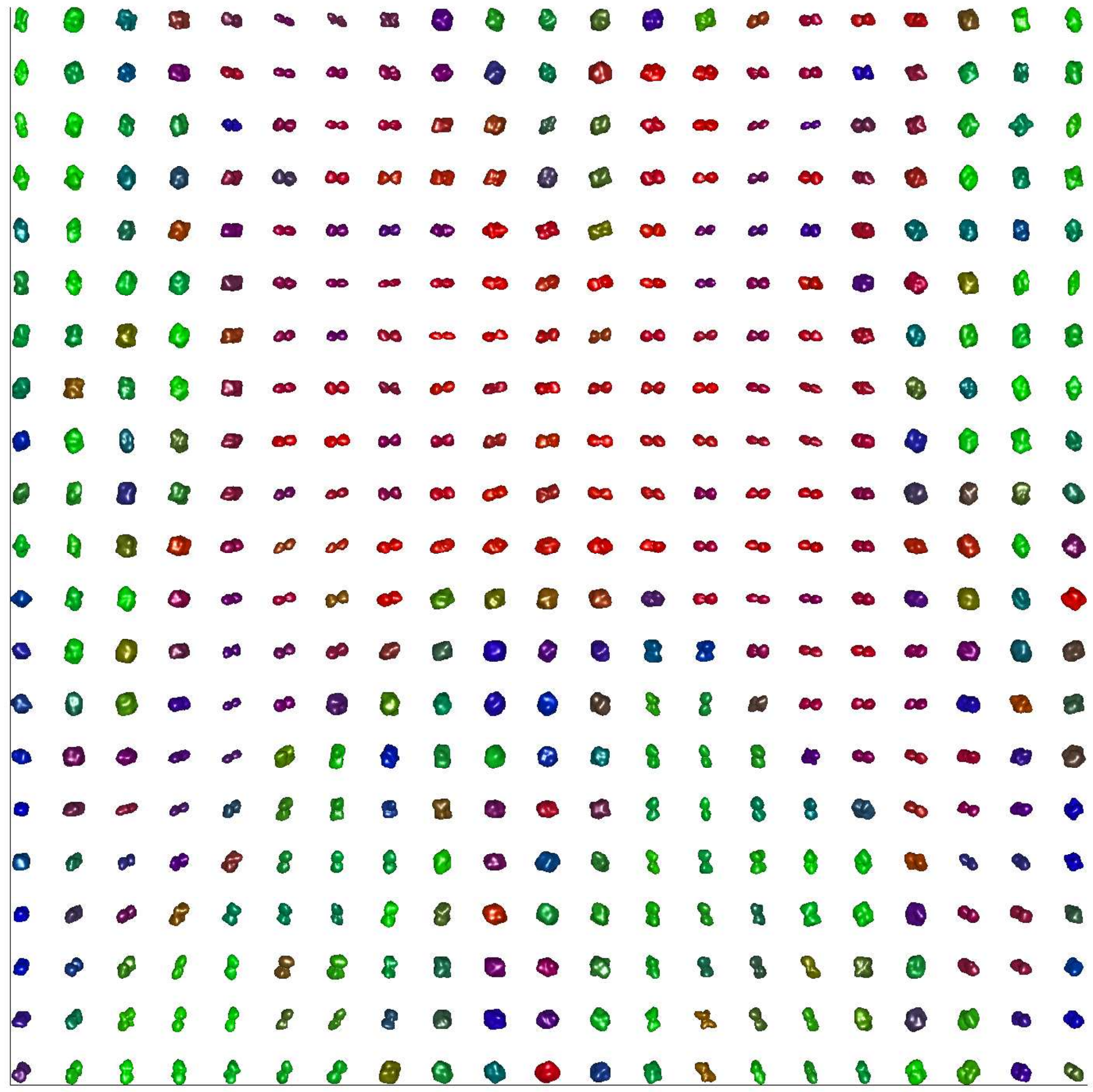}}
	\subfigure[]{\includegraphics[scale=0.145]{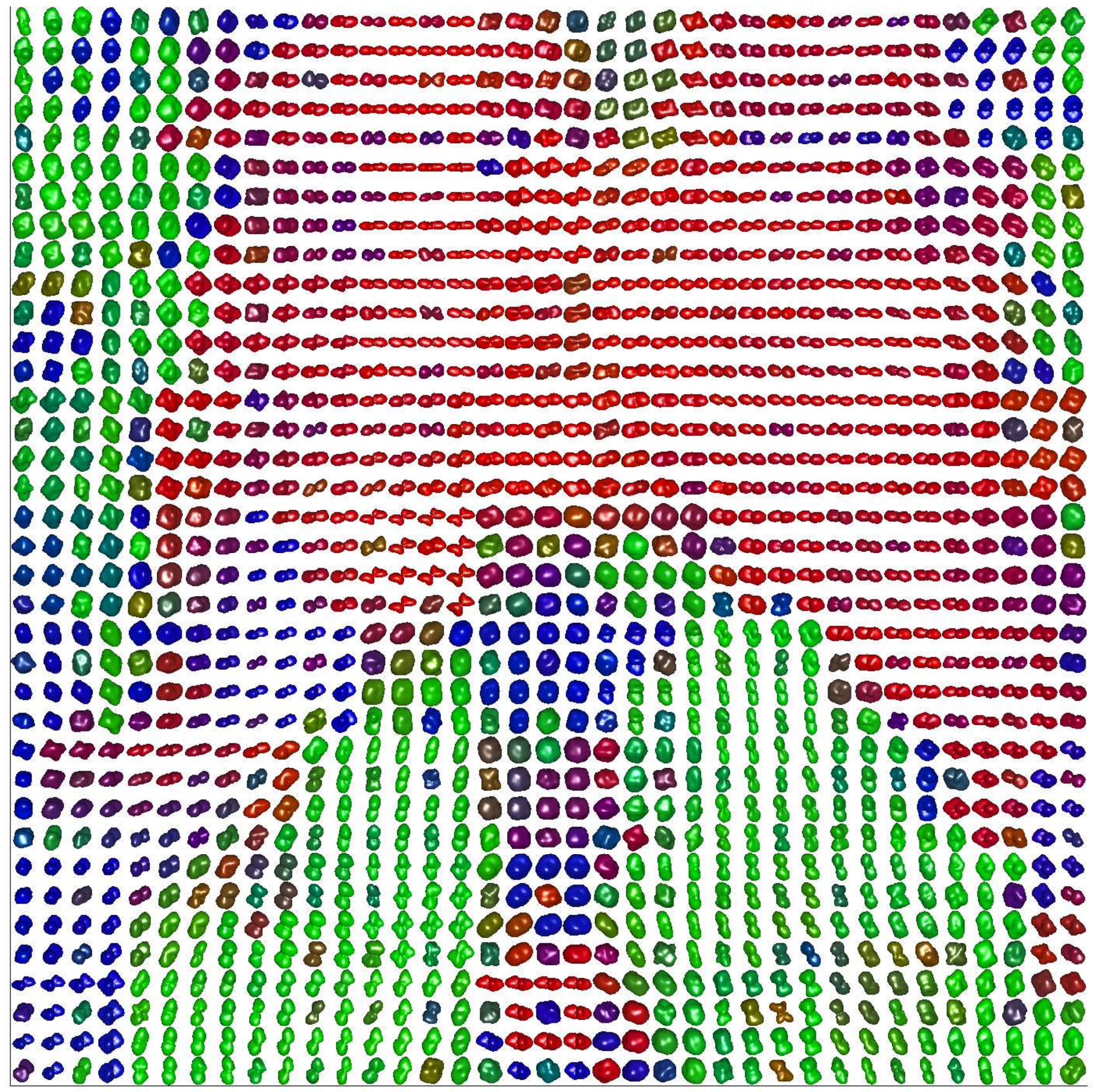}}
	\subfigure[]{\includegraphics[scale=0.145]{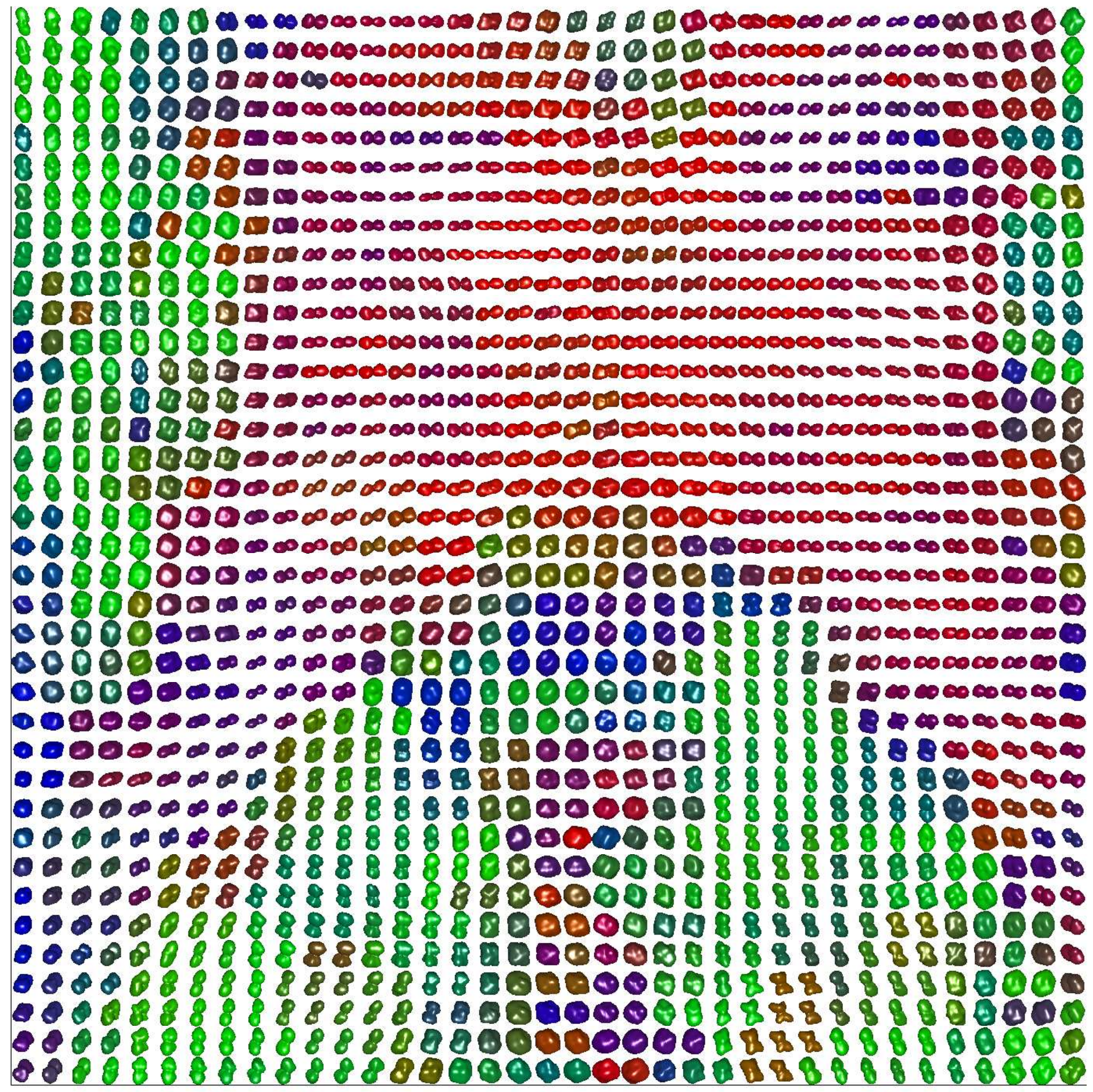}}	
	\caption{Graphic results for interpolation of rank-6 real HOT field: (a) Ground-truth, (b) Training data, (c) linear
		interpolation, (d) Tucker decomposition process.}
	\label{real6}
\end{figure}

Figures \ref{real4} and \ref{real6} and Table \ref{tab2} show
qualitative and quantitative results for HOT interpolation in rank-4
and 6 synthetic data. The TDP improves the performance when compared
to linear interpolation.

\begin{table}[ht!]
	\centering
	\caption{Frobenius distance for rank-4 and 6 real HOT fields}
	\begin{tabular}{ccc}
		\hline
		& Rank-4 & Rank-6 \\
		\hline
		TDP   & $2.030\pm1.124$ & $1.947\pm1.573$ \\
		Direct interpolation & $2.727\pm2.601$ & $2.664\pm2.334$ \\
		\hline
	\end{tabular}%
	\label{tab2}%
\end{table}%

\section{Discussion and Conclusions}
The Tucker decomposition process demonstrates better performance in
the interpolation of HOT fields of any order, compared to direct
linear interpolation. In rank-2 data, the TDP also outperforms
log-Euclidean interpolation and obtains a similar result to a recently
proposed framework based on generalized Wishart processes. Our method
can capture the global spatial trend of the field.  For this reason,
it can deliver a precise estimation of new data. The TDP is flexible
enough to model several transitions inside HOT fields. This property
is important because it allows the TDP to adapt to heterogeneous HOT
data.

Qualitative results of figures \ref{st2}, \ref{st4}, \ref{st6},
\ref{real2}, \ref{real4}, \ref{real6} illustrate an interesting
behavior when there are strong changes among nearby tensors. Looking
the figures in detail, the direct interpolation can not capture this
rapid transitions inthe field, no matter the rank. Although the
interpolation performed by TDP is not very precise for these cases, it
is more accurate than the linear method.

Due to the smooth nature of the GPs appearing in the matrix
$\boldA(\boldz)$, the TDP can not follow abrupt changes in data
(i.e. crossing fibers, outliers and transitions from red to green
tensors). Therefore, the error in the prediction increases for these
particular cases. A potential alternative is to use GPs with a
non-stationary covariance function. For example, to assume that the
length-scale hyper-parameter of the kernel is a function of an
indexation variable (i.e. spatial coordinates \textbf{z}).

\section*{Acknowledgments}
 H.D. Vargas Cardona is funded by Colciencias under the program:
 \textit{formaci\'on de alto nivel para la ciencia, la tecnolog\'ia y
   la innovaci\'on - Convocatoria 617 de 2013.}

\bibliographystyle{plain}
\bibliography{biblio}

\end{document}